%% file: main.tex
\documentclass[11pt]{article}
\usepackage[margin=1.1in]{geometry}
\input{packages_and_commands.tex}
\usepackage[numbers, compress]{natbib}
\usepackage{hyperref}
\usepackage{url}
\usepackage{microtype}
\usepackage{graphicx}
\usepackage{booktabs} %
\usepackage{hyperref}

\usepackage{amsmath}
\usepackage{amssymb}
\usepackage{mathtools}
\usepackage{amsthm}
\usepackage{etoc}

\title{Watermark Smoothing Attacks in Language Models}
\date{}
\author{Hongyan Chang \\
National University of Singapore \\
\texttt{hongyan@comp.nus.edu} 
\and Hamed Hassani \\
University of Pennsylvania \\
\texttt{hassani@seas.upenn.edu} \\
\and Reza Shokri \\
National University of Singapore \\
\texttt{reze@comp.nus.edu}
}

\begin{document}

\maketitle

\input{sec/abstract}

\input{sec/intro}
\input{sec/background}

\input{sec/story}
\input{sec/attack}

\input{sec/exp}

\input{sec/conclusion}

\bibliographystyle{IEEEtranN}
\bibliography{references}

\clearpage
\appendix
\addcontentsline{toc}{part}{Appendix}
\part*{Appendix}
\etocsetnexttocdepth{2} 
\localtableofcontents

\include{data/tab_notation.tex}
\include{sec/appendix_additional_results.tex}

\include{sec/appendix_analysis.tex}
\include{sec/appendix_limitations.tex}
\end{document}

%% file: packages_and_commands.tex
\usepackage{amsmath,amssymb,amsfonts} %
\usepackage{graphicx} %
\usepackage{textcomp} %

\usepackage{booktabs} %
\usepackage{tabularx} %
\usepackage{pgfplots} %
\usepackage{pgfplotstable} %
\usepgfplotslibrary{fillbetween} %
\usetikzlibrary{patterns}
\usetikzlibrary{positioning}

\usepackage{multirow} %
\usepackage{algorithm} %
\usepackage{algorithmic} %
\renewcommand{\algorithmicforall}{\textbf{for each}} %
\usepackage{xr} %
\usepackage{caption} %
\usepackage{tikz} %
\tikzstyle{roccurvestyle}=[mark=none] %
\usepackage{mathrsfs} %

\usepackage{etoolbox} %
\AtBeginEnvironment{tabular}{\footnotesize} %

\usepackage{hyperref} %
\usepackage{adjustbox} %
\usepackage{wrapfig} %
\usepackage{bbm} %

\usepackage[makeroom]{cancel} %
\usepackage{array} %
\usepackage[capitalize]{cleveref} %

\usepackage{xcolor} %

\definecolor{darkblue}{rgb}{0.157, 0.267, 0.447}
\definecolor{textblue}{RGB}{0,0,255}
\definecolor{blue}{RGB}{31,120,180}
\definecolor{green}{RGB}{51,160,44}
\definecolor{red}{RGB}{226,129,128}
\definecolor{bblue}{RGB}{166,206,227}
\definecolor{bgreen}{RGB}{178,223,138}
\definecolor{bred}{RGB}{253,200,200}
\definecolor{mygray}{gray}{0.6}
\definecolor{oblue}{RGB}{0,0,255}
\definecolor{ored}{RGB}{255,0,0}
\definecolor{tiger}{HTML}{FC6A03}

\usepackage{xcolor}
\usepackage{subcaption}    %

\pgfplotsset{
    box plot/.style={
        /pgfplots/.cd,
        only marks,
        mark=-,
        mark size=0.45em,
        /pgfplots/error bars/.cd,
        y dir=plus,
        y explicit,
    },
    box plot box/.style={
        /pgfplots/error bars/draw error bar/.code 2 args={%
            \draw  ##1 -- ++(0.4em,0pt) |- ##2 -- ++(-0.4em,0pt) |- ##1 -- cycle;
        },
        /pgfplots/table/.cd,
        y index=2,
        y error expr={\thisrowno{3}-\thisrowno{2}},
        /pgfplots/box plot
    },
    box plot top whisker/.style={
        /pgfplots/error bars/draw error bar/.code 2 args={%
            \pgfkeysgetvalue{/pgfplots/error bars/error mark}%
            {\pgfplotserrorbarsmark}%
            \pgfkeysgetvalue{/pgfplots/error bars/error mark options}%
            {\pgfplotserrorbarsmarkopts}%
            \path ##1 -- ##2;
        },
        /pgfplots/table/.cd,
        y index=4,
        y error expr={\thisrowno{2}-\thisrowno{4}},
        /pgfplots/box plot
    },
    box plot bottom whisker/.style={
        /pgfplots/error bars/draw error bar/.code 2 args={%
            \pgfkeysgetvalue{/pgfplots/error bars/error mark}%
            {\pgfplotserrorbarsmark}%
            \pgfkeysgetvalue{/pgfplots/error bars/error mark options}%
            {\pgfplotserrorbarsmarkopts}%
            \path ##1 -- ##2;
        },
        /pgfplots/table/.cd,
        y index=5,
        y error expr={\thisrowno{3}-\thisrowno{5}},
        /pgfplots/box plot
    },
    box plot median/.style={
        /pgfplots/box plot
    }
}

\definecolor{DarkBlue}{RGB}{0,58,125}    %
\definecolor{MedBlue}{RGB}{0,141,255}    %
\definecolor{Pink}{RGB}{255,115,182}     %
\definecolor{Purple}{RGB}{199,1,255}     %
\definecolor{Green}{RGB}{78,203,141}     %
\definecolor{Orange}{RGB}{255,157,58}    %
\definecolor{Yellow}{RGB}{249,232,88}    %
\definecolor{Red}{RGB}{216,48,52}        %

%% file: sec/abstract.tex
\begin{abstract}

Watermarking is a key technique for detecting AI-generated text. In this work, we study its vulnerabilities and introduce the \textit{Smoothing Attack}, a novel watermark removal method. By leveraging the relationship between the model's confidence and watermark detectability, our attack selectively smoothes the watermarked content, erasing watermark traces while preserving text quality. We validate our attack on open-source models ranging from $1.3$B to $30$B parameters on $10$ different watermarks, demonstrating its effectiveness. Our findings expose critical weaknesses in existing watermarking schemes and highlight the need for stronger defenses.

\end{abstract}

%% file: sec/intro.tex
\section{Introduction}
Detecting whether a text is generated by language models is critical in domains like fraud detection, fake news identification, and plagiarism prevention. A common approach is watermarking, where subtle patterns are embedded in the generated text for later detection~\cite{aaronson2023watermarking, christ2023undetectable, huang2023towards, li2024statistical}. Watermarking has gained traction in both academia and industry~\cite{dathathri2024scalable} as a key safeguard for language model applications. While various watermarking techniques exist, they share a core principle: favoring certain tokens over others (detailed in Section~\ref{sec:background}).

In this work, we identify key scenarios where watermarks fail and introduce a novel watermark removal attack that exploits this weakness, revealing fundamental limitations in existing watermarking schemes.

\paragraph{Effectiveness of watermarks.} We say a watermark is effective if \textbf{(i)} the watermarked text maintains high quality, comparable to those generated from the corresponding un-watermarked model, and \textbf{(ii)} the detector reliably identifies watermark traces, i.e., it can identify watermarked text without making a large error. We analytically and empirically show that these aspects are in tension: better text quality often implies lower watermark detectability, and vice versa. Moreover, both are connected through the model's confidence in generating output. We explain the high-level idea as follows (see more detail in Section~\ref{sec:limit}).

Given a prefix, when the model is confident about the output token, watermarking has negligible impact on the output. In this case, the watermark trace is not obvious. Conversely, when the model is not confident, watermarking makes the model tend to select certain tokens (that are originally unlikely to get sampled) over others, making watermark trace more detectable while degrading the text quality. 

\paragraph{Smoothing Attack.} Leveraging this insight, we propose the \textit{Smoothing Attack} for watermark removal. For each prefix, the attack first identifies if the output token contains the watermark trace, by estimating the target watermarked model's confidence in this output. If the confidence is low, then we replace the token with a freshly sampled one (see more detail in Section~\ref{sec:smoothing-attack}), removing watermark traces while maintaining text quality; otherwise, if the confidence is high, then we retain the watermarked model’s output. 

We evaluate our attack across ten diverse watermarking schemes and three different families of open-sourced models, OPT~\cite{zhang2022opt} (from 1.3B to 30B parameters), Llama3-8B~\cite{dubey2024llama} and Qwen2-1.5B~\cite{chu2024qwen2}.
In certain cases, our attack completely removes the watermark (reducing watermark detection rates to zero) while preserving the text quality. Our attack can also outperform the state-of-the-art \textit{Paraphrasing Attack}, which uses the strong GPT-3.5-turbo to paraphrase the watermarked text. Compared with \textit{Paraphrasing Attack}, our attack is more cost-efficient, as it uses only much weaker reference models, e.g., OPT-125M~\citep{zhang2022opt} when attacking OPT models from $1.3$B to $30$B parameters. These findings underscore critical weaknesses in existing watermarks and highlight the need for more robust defenses.

%% file: sec/background.tex
\section{Preliminaries and Related Work}\label{sec:background}

For an auto-regressive language model (LM) $M$, we use $\mathcal{V}$ to denote its vocabulary (i.e., the set of all possible tokens). On a given prompt, $M$ generates its output as follows. At each token position $t$ with the given prefix (including the previously generated tokens and the prompt), model $M$ first computes the logit for each token $v$, written as $l_t(v)$. Applying the soft-max function to the logits, we obtain the following probability distribution for the output token.
\begin{align}\label{eq:soft-max}
	P_t(v) &=
\frac{\exp\bigl(l_t(v)\bigr)}{\sum_{v' \,\in\, \mathcal{V}} \exp\bigl(l_t(v')\bigr)}.
\end{align}

With $P_t$, two sampling strategies are often employed to sample the next token. Top-k sampling~\cite{fan2018hierarchical, holtzman2018learning} samples a token from the $k$ most probable candidates. Top-p/Nucleus sampling~\cite{holtzman2019curious} samples a token from the smallest set of tokens whose cumulative sum of probability masses exceeds some constant $p$. We denote the output text, i.e., a sequence of tokens, as $(v_1, \ldots, v_T)$ for some positive $T$.

At a high level, watermarks are embedded into the generated text $(v_1, \ldots, v_T)$ through specific patterns of tokens. The detector tries to find traces of such patterns, by computing some detection score function $d(v_1, \ldots, v_T)$. If the score exceeds a certain threshold $\tau$, then the text is predicted as generated from the watermarked model. Next, we explain representative watermarking algorithms.

\paragraph{Green-red list watermark~\cite{kirchenbauer23a}.} The idea is to modify the logits of specific tokens. In particular, at each token position $t$, the vocabulary set $V$ is partitioned to the red and green lists, where the green list, denoted as $\mathcal{G}_t$, takes a $\gamma$ fraction of the vocabulary. Logits of tokens in $\mathcal{G}_t$ are increased by some pre-fixed constant $\delta$; while the logits of other tokens remain unchanged. The modified probability distribution is written as
\begin{align}\label{eq:modified_p_green_list}
\widetilde{P}_t(v) 
\;&=\;
\frac{\exp\bigl(l_t(v) + \delta \cdot \mathbf{1}{\{v \in \mathcal{G}_t\}}\bigr)}%
     {\sum_{v' \,\in\, \mathcal{V}} \exp\Bigl(l_t(v') + \delta \cdot \mathbf{1}{\{v' \in \mathcal{G}_t\}}\Bigr)}.
\end{align}
A sampling strategy, either top-k or top-p, is then applied to $\widetilde{P}_t$ to output the next token. The detector looks for evidence that the green tokens appear disproportionally more frequently. Accordingly, given $(v_1, ..., v_T)$ and the green lists $\{\mathcal{G}_t\}_{t=1}^T$ , the detector computes 
\begin{align}\label{eq:detection_score_green}
d(v_1, \ldots, v_T) = \frac{ \sum_{t=1}^T \left(\mathbf{1}{\{v_t \in \mathcal{G}_t\}} -\gamma \right)}{\sqrt{T \gamma (1 - \gamma)}}, 
\end{align}
and then predicts the given text as watermarked if this score exceeds some threshold $\tau$. Here $\gamma$ stands for the expected number of green tokens per token position generated by any other models, since the \textit{assignment of the green list} $\mathcal{G}_t$ is random and is known to the LM provider and the detector only. The denominator $\sqrt{T \gamma (1 - \gamma)}$ normalizes the detection score: it is unlikely that a non-watermarked text will be misclassified as watermarked, particularly when the text is long enough (i.e., $T$ is large). 

\paragraph{Gumbel sampling watermark~\cite{kuditipudi2023robust,aaronson2023watermarking}.} The idea is to use Gumbel sampling~\cite{gumbel1954statistical} when sample the token at each position $t$. In particular, they first sample $u_t(v)$ from $[0,1]$ for each $v$, based on some random seed computed from the preceding $k$ tokens ($k$ is some hyperparameter) and some \textit{secret watermarking key}. The output token $v_t^*$ is then selected based on the sampled outcomes and the original $P_t(v)$, given as 
$v_t^* = \arg\max_{v \in \mathcal{V}} - \frac{\log u_t(v)}{P_t(v)}.$
The watermarked text tends to contain token $v$ that is associated with a larger $u_t(v)$. The detection score is computed as 
$d(v_1,\ldots,v_T)=- \sum_{t=1}^T \log(1 - u_t(v_t))$. If this score exceeds some threshold, the text is predicted as watermarked.

\paragraph{Tournament sampling watermark~\cite{dathathri2024scalable}.}
The idea is to use an additional tournament process to select the output token. Specifically, when generating the $t$-th token, $m$ random watermarking functions $g^{(1)}..., g^{(m)}$ are used to assign $m$ scores for each token in the vocabulary. The scores depend on the given token $v$ and a random seed $r_t$ computed from the secret watermarking key and the recent context (e.g., preceding $k$ tokens), and are denoted as $g^{(l)}(v,r_t)\in\{0,1\}$. With the scores, the next token is selected as follows. First, $2^m$ tokens are sampled with replacement from the original probability distribution $P_t$. These sampled $2^m$ tokens are then split into $2^{m-1}$ pairs of competing tokens and the tokens with larger scores win (breaking ties randomly). This process is repeated for $m$ times and the final winner is the output token. As this process favors tokens with larger tournament scores, the detector computes $
d(v_1,\ldots, v_T) = \frac{1}{T} \sum_{t=1}^T \frac{1}{m}\sum_{l=1}^m g^{(l)}(v_t, r_t),$
and predicts the given text as watermarked if this score exceeds $0.5$ by a large margin. Dathathri et al.~\cite{dathathri2024scalable} also provide an equivalent way to sample the token, using some modified proability distribution $\widetilde P_t$ computed from the $m$ watermarking functions and the original $P_t$ (detailed in Appendix~\ref{appendix:related}).

\paragraph{Other related work.} 
\input{sec/related_work}

%% file: sec/related_work.tex
The Green-red list watermark~\cite{kirchenbauer23a} favors tokens in the green lists and introduces distortions to the distribution of the output tokens. Therefore, we often say it is a \textit{distortionary} watermark. Variations of this scheme mainly differ in the assignment of green lists and the detection process, e.g., see~\cite{kirchenbauer2023reliability,lee2023wrote,liu2023unforgeable,wu2023dipmark}. Gumbel and Tournament sampling watermarks~\cite{kuditipudi2023robust,aaronson2023watermarking,dathathri2024scalable} are distortion-free, preserving the original model's token distribution (when averaging over all possible secret watermarking keys). We refer readers to the original papers for detailed analyses. There are also other distortion-free watermarks, e.g., see \cite{huunbiased,christ2023undetectable,fairoze2023publicly,christ2024pseudorandom,ghentiyala2024new,golowichedit,wu2023dipmark}. We refer to~\cite{zhao2024sok} for an in-depth survey. In our evaluation, we evaluate $10$ representative watermarks, spanning distortionary and distortion-free approaches, to demonstrate the universal applicability of our attack. 

The canonical way to remove watermark is by disrupting its patterns, via injecting special characters, homoglyphs, or emojis into the text \citep{gabrilovich2002homograph,helfrich2012dual,pajola2021fall,boucher2022bad,goodside2023adversarial}. However, such modifications often reduce the text quality significantly. Another strategy is to paraphrase the watermarked text using another model \cite{kirchenbauer2023reliability,sadasivan2023can,krishna2023paraphrasing,piet2023mark,jovanovic2024watermark,zhangwatermarks2024}, referred to as \textit{Paraphrasing Attacks}. The performance of such an attack typically relies on the power of the model it uses. A large LLM often leads to good quality in the paraphrased texts, e.g., using the GPT-3.5-turbo \cite{openai2023gpt} to paraphrase the watermarked texts generated from the 7B-parameter Llama \cite{touvron2023llama}.

%% file: sec/story.tex
\section{On the Effectiveness of Watermarks}\label{sec:limit}

In this section, we investigate the key factors that contribute to the effectiveness of watermarking algorithms, namely watermark detectability and text quality. Our key finding is that these two factors are in tension: better text quality often implies lower watermark detectability, and vice versa. Moreover, both are tightly connected to the model's confidence in generating output, which reveals a vulnerability exploitable by the attack we propose later. The detailed derivations are in Appendix~\ref{appendix:analytical_analysis}.

\begin{figure*}[t]
\centering
\input{figure_scripts/fig_scatter_pg_delta_KGW.tex}
\caption{The correlations among $S_t$ (watermark contribution score), $\mathbb{E}_{v \sim P_t}\left[\mathbf{1}\{v \in \mathcal{G}_t\}\right]$ (expected number of green tokens from the un-watermarked model), and $\|P_t\|^2$ (model confidence), evaluated on model OPT-1.3B with the Red-green list watermark with $\gamma=0.5$ and $\delta=1.0$. The values are computed from different prefixes, constructed from the text from the Wikipedia article about Harry Potter.}
\label{fig:kwg_l2_norm}
\end{figure*}
\begin{figure*}[thbp]
\centering
\input{figure_scripts/fig_scatter_pg_delta_gumble_synthID.tex}
\caption{
The correlation between $S_t$ (watermark contribution score) and $\|P_t\|^2$ (model confidence) evaluated on model OPT-1.3B with the {Gumbel} and {Tournament sampling} (with $m$ tournaments) watermarks, using the same setup as in Figure~\ref{fig:kwg_l2_norm}. Each sample corresponds to a specific prefix and secret key. $\|P_t\|^2$ is computed from the original un-watermarked model. The overall observation is similar to what we have for the \textit{Green-red list watermarking}: $S_t$ decreases as $\|P_t\|^2$ increases.}
\label{fig:synthID_gumbel_l2_norm}
\end{figure*}

\subsection{From watermark detection to model's confidence}\label{sec:detection-confidence}

Recall that the watermark detectability, which is characterized by the detection score (e.g., see Eq.~\eqref{eq:detection_score_green}), depends on all tokens in the given text. Our key finding is that the individual contributions to the detection score from tokens at different positions depend on the model prediction confidence. Thus, to estimate this token-level contribution to watermark detectability, it suffices to estimate the model's confidence. 

In what follows, we go through the reasoning using the Green-red list watermark~\cite{kirchenbauer23a} as an illustration. We then show that our findings generalize to the other two representative watermarks, Tournament sampling and Gumbel sampling~\cite{kuditipudi2023robust,aaronson2023watermarking,dathathri2024scalable}.

\paragraph{Token-level contribution to watermark detection.} Recall Eq.~\eqref{eq:detection_score_green}, the detection score (roughly) counts the number of tokens that belong to the green list in all token positions. For each token position $t$ and the assigned green list $\mathcal{G}_t$,  we define  
\begin{align}\label{eq:contribution-green}
S_t = \mathbb{E}_{v \sim \widetilde{P}_t}\left[\mathbf{1}\{v \in \mathcal{G}_t\}\right] - \mathbb{E}_{v \sim P_t}\left[\mathbf{1}\{v \in \mathcal{G}_t\}\right],
\end{align}
where $P_t$ and $\widetilde P_t$ stand for the original probability distribution and the modified one (all logits of green tokens are increased by some $\delta$), respectively. 

Comparing Eq.~\eqref{eq:contribution-green} with the detection score in Eq.~\eqref{eq:detection_score_green}, we have omitted the normalization factor $\sqrt{T\gamma(1-\gamma)}$ for brevity. We also focused on one particular position $t$ and the corresponding assignment of $\mathcal{G}_t$. If we take the expectation over the all possible assignments of $\mathcal{G}_t$, then the subtracted term $\mathbb{E}_{v \sim P_t}\left[\mathbf{1}\{v \in \mathcal{G}_t\}\right]$ becomes $\gamma$, since $\mathcal{G}_t$ takes a $\gamma$ fraction of tokens from the vocabulary. 

$S_t$ captures the increment of the token-level contribution to the overall detection score due to watermarking, as the first expectation is taken over $\widetilde P_t$ rather than $P_t$. Overall, larger $S_t$ leads to higher watermark detectability, and vice versa. 

\paragraph{Connecting $S_t$ to model's confidence.} Recall that $\delta$ represents the shift applied to the logits of tokens in the green list. Then we can write $S_t$ as
\begin{align}\label{eq:green-function-analysis}
S_t = \frac{-(e^\delta - 1)\cdot \mathbb{E}_{v \sim P_t}\left[\mathbf{1}\{v \in \mathcal{G}_t\}\right] + (e^\delta - 1)}{ 1/{\mathbb{E}_{v \sim P_t}\left[\mathbf{1}\{v \in \mathcal{G}_t\}\right]}+(e^\delta - 1)},
\end{align}
which is a function of $\mathbb{E}_{v \sim P_t}\left[\mathbf{1}\{v \in \mathcal{G}_t\}\right]$. Clearly, $S_t$ can be different at different token positions. 

To validate Eq~\eqref{eq:green-function-analysis}, we draw $400$ sample points on $S_t$ and $\mathbb{E}_{v \sim P_t}\left[\mathbf{1}\{v \in \mathcal{G}_t\}\right]$ using the OPT-1.3B model with  $\gamma=0.5$ and $\delta=1.0$ and plot them in Figure~\ref{fig:kwg_l2_norm} (the left-most subfigure). The empirical observations align with our analysis.

Further, we show that $\mathbb{E}_{v \sim P_t}\left[\mathbf{1}\{v \in \mathcal{G}_t\}\right]$ is correlated with the squared $\mathcal{L}_2$ norm of the probability vector $P_t$ at position $t$, denoted as $\|P_t\|^2$. In particular, the mean of $\mathbb{E}_{v \sim P_t}\left[\mathbf{1}\{v \in \mathcal{G}_t\}\right]$ is $\gamma$ and the variance is $\gamma(1-\gamma)\|P_t\|^2$. When $\|P_t\|^2=1$,  $\mathbb{E}_{v \sim P_t}\left[\mathbf{1}\{v \in \mathcal{G}_t\}\right]$ is either $0$ or $1$. Conversely, when $\|P_t\|^2=1/|\mathcal{V}|$, $\mathbb{E}_{v \sim P_t}\left[\mathbf{1}\{v \in \mathcal{G}_t\}\right]$ is concentrated around $\gamma$, a value smaller than $1$. We demonstrate this correlation in~\autoref{fig:kwg_l2_norm} (the middle subfigure), using the same $400$ sample points obtained above. 

The value of $\|P_t\|^2$ measures how confident the model is when outputting the next token at position $t$. In particular, $\|P_t\|^2$ attains its maximum value $1$, i.e., all probability masses are concentrated on a single token when the model is absolutely certain of its output; and $\|P_t\|^2$  attains its minimum value $1/|\mathcal{V}|$, i.e., the probability masses are evenly distributed over all tokens, when the model has no idea which token to output.

Putting everything together, we are able to connect $S_t$ to the model's confidence, which is characterized by $\|P_t\|^2$. We demonstrate this correlation in~\autoref{fig:kwg_l2_norm} (the right subfigure). We observe that when $\|P_t\|^2$ is large (in particular, consider $\|P_t\|^2=1$), then $\mathbb{E}_{v \sim P_t}\left[\mathbf{1}\{v \in \mathcal{G}_t\}\right]$ is close to $0$ or $1$, leading to a relatively small value of $S_t$. Conversely, when $\|P_t\|^2$ is small (in particular, consider $\|P_t\|^2=\frac{1}{|\mathcal{V}|}$), then $\mathbb{E}_{v \sim P_t}\left[\mathbf{1}\{v \in \mathcal{G}_t\}\right]$ is roughly $\gamma$, a value smaller than $1$, leading to a relatively large value of $S_t$. 

In summary, \textit{when the model is more confident in choosing the output token at position $t$, then its contribution $S_t$ to the watermark detectability is smaller, and vice versa.} 

\paragraph{Generalization to other watermarking solutions.} For Gumbel sampling, we define the token-level contribution to watermark detection as
$S_t = - \log (1 - U_{v^*}) -  \mathbb{E}_{v \sim P_t}[- \log(1 - U_{v})],$
where $v^*$ is the token selected by the watermarked model. Note that the choice of $v^*$ is deterministic after the secret key held by the LM provider and the prefix content are fixed. For Tournament sampling, we define the token-level contribution as
$S_t = \mathbb{E}_{v \sim \widetilde{P}_t}\left[\frac{1}{m}\sum_{l=1}^m g^{(l)}(v,r) \right] - \mathbb{E}_{v \sim P_t}\left[\frac{1}{m}\sum_{l=1}^m g^{(l)}(v,r) \right]$, where $\widetilde{P}_t$ is the modified probability distribution. 

For these two watermarks, we still observe the same correlation between $S_t$ and $\|P_t\|^2$ as we have for Green-list watermarks, as shown in \autoref{fig:synthID_gumbel_l2_norm}. Namely, the token-level contribution $S_t$ to the watermark detectability is negatively correlated to the model's confidence at position $t$.

\begin{figure*}[t]
\centering
\input{figure_scripts/fig_scatter_l2_dtv.tex}
\caption{The correlation between $D_{TV}(P_t,\widetilde P_t)$, i.e., the negative impact on text quality due to watermarks (in color blue), and $\|P_t\|^2$ measured on the OPT-1.3B with the Green-red list, Gumbel sampling, and Tournament sampling watermarks. We also plot $D_{TV}(P_t,P^{\text{ref}}_t)$, which measures the negative impact on text quality if we alternatively sample from the reference model OPT-125M (in color red). }
\label{fig:quality_all}
\end{figure*}
\begin{figure*}[t]
\centering
\input{figure_scripts/fig_scatter_st_dtv.tex}
\caption{The correlation between $D_{TV}(P_t,\widetilde P_t)$, i.e., the negative impact of watermarking on the text quality, and $S_t$, i.e., the token-level contribution to watermark detectability. We measure the correlation on the OPT-1.3B model. For all three watermarking schemes, $D_{TV}(P_t,\widetilde P_t)$ increases as $S_t$ increases. }
\label{fig:quality_contribution}
\end{figure*}

\subsection{Impact of watermarking on text quality}

Next, we show that the impact of watermarking on the text quality also depends on the model's confidence in generating its output token.

\paragraph{Total variation distance (TVD).} To measure the impact of watermarking on the text quality at any token position $t$, we use the total variation distance between the probability distributions of the original un-watermarked model and the watermarked model, defined as $
D_{TV}(P_t, \widetilde P_t) = \frac{1}{2} \sum_{v \in \mathcal{V}} |P_t(v) - \widetilde P_t(v)|,    $
where $P_t$ corresponds to the original probability distribution and $\widetilde P_t$ is the watermarked one. 

Due to the large vocabulary size, we measure $d_{TV}(P_t, \widetilde P_t)$ empirically, by repeatedly sampling $100$ tokens from each distribution and computing the variation between the sampled tokens' frequencies. We repeat this process for different token positions, with the prefixes constructed from the Wikipedia article about Harry Potter~\footnote{\url{https://en.wikipedia.org/wiki/Harry_Potter_(film_series)}}, similar to the setup in Section~\ref{sec:detection-confidence}. A small $D_{TV}(P_t, \widetilde P_t)$ indicates that the watermark introduces a small distortion to the original probability distribution at position $t$, suggesting a negligible impact on the model's output quality; and vice versa. 

\paragraph{TVD depends on model's confidence. }The results for watermarked texts are shown in~\autoref{fig:quality_all}.  (in color blue). For all three watermarking schemes, $D_{TV}(P_t,\widetilde P_t)$ decreases as $\|P_t\|^2$ increases, meaning that the negative impact on text quality due to watermarks decreases as the model becomes more confident in its output. Conversely, when the model is not confident in its output, the negative impact of watermarks on the text quality becomes more notable. This is because, when the model is less confident, watermarks are able to increase the probabilities of certain tokens more significantly than the cases when the model is more confident (in such cases, watermarks do not make a notable difference in the model's output). 

We also plot $D_{TV}(P_t,P^{\text{ref}}_t)$, which measures the negative impact on text quality if we alternatively sample from the reference model OPT-125M (in color red). We note that when the model is not confident in its output, i.e., when $\|P_t\|^2$ is small, sampling from the reference model's token distribution, i.e., $P^{\textrm{ref}}_t$, does not hurt the text quality. In particular, under the Green-red list watermarking scheme, $D_{TV}(P_t, P^{\textrm{ref}}_t)$ is comparable to  $D_{TV}(P_t, \widetilde P_t)$ when $\|P_t\|^2$ is small (observe that the red points generally overlap with the blue ones). For Gumbel and Tournament sampling, $D_{TV}(P_t, P^{\textrm{ref}}_t)$ is even smaller than $D_{TV}(P_t, \widetilde P_t)$ when $\|P_t\|^2$ is small (observe that the red points are generally below the blue ones). Conversely, when the model is confident in its output, i.e., when $\|P_t\|^2$ is large, replacing the watermarked model with a reference model may hurt the text quality (observe that the red points are above the blue ones). 

In summary, the impact of watermarks on the text quality depends on the model's confidence. When the model's confidence is high, watermarking has a negligible impact on the text quality. Conversely, when the model's confidence is low, watermarking has a notable negative impact on the text quality. Surprisingly, in this case, we can even replace the watermarked model with a much smaller reference model, achieving comparable and even better text quality.

\paragraph{The limitations of existing watermarks.} We come to realize that the two aspects of the watermark effectiveness --- watermark detectability and text quality --- are correlated with the model's confidence in its output. That means these two aspects are also interconnected. In Figure~\ref{fig:quality_contribution}, we plot the correlation between $D_{TV}(P_t,\widetilde P_t)$ and $S_t$, empirically measured on OPT-1.3B model using the same setup as the above simulations. When the watermark has little impact on text quality (i.e., smaller total variation distance), the watermark is also less detectable (i.e., smaller $S_t$). Conversely, tokens that contribute more to watermark detection also lead to more notable text quality degradation. 

This finding, in turn, reveals the crucial limitation of existing watermarking schemes: high watermark detectability and high text quality cannot be achieved at the same time, since the very same set of tokens causes quality degradation while contributing to watermark detectability simultaneously. 

%% file: figure_scripts/fig_scatter_pg_delta_KGW.tex
\begin{tikzpicture}[]
    \begin{axis}[scale=1, width=0.32\linewidth,height=0.2\linewidth, ymin=0,ymax=0.5,yticklabel= {\pgfmathprintnumber\tick}, xlabel style={yshift=0.5em},ylabel={\small $S_t$},ylabel near ticks, enlarge x limits=0.05, xticklabel= {\pgfmathprintnumber\tick}, xlabel={\small \( \mathbb{E}_{v \sim P_t}\left[\mathbf{1}\{v \in \mathcal{G}_t\}\right] \)}, x tick label style={font=\small,text width=1cm,xshift=0.35cm}, enlarge y limits=0.1, y tick label style={font=\small}, xmin=0,xmax=1,xtick={0, 0.2, 0.4, 0.6, 0.8,1.0},legend pos=north east,legend cell align=left,legend style={draw},legend style={font=\scriptsize},axis background/.style={}]
        \addplot[blue,only marks,mark size=1.2pt] table[x=PgUnwatermarked, y=Delta] {data/illustration/KGW_OPT_1_3b/all.dat};
        
    \end{axis}
\end{tikzpicture}%
\begin{tikzpicture}[]
    \begin{axis}[scale=1,width=0.32\linewidth,height=0.2\linewidth, xlabel style={yshift=0.5em}, xmin=0,xmax=1.0,yticklabel= {\pgfmathprintnumber\tick}, ylabel={\small \( \mathbb{E}_{v \sim P_t}\left[\mathbf{1}\{v \in \mathcal{G}_t\}\right] \)},ylabel near ticks, enlarge x limits=0.05, xticklabel= {\pgfmathprintnumber\tick}, xlabel={\small $\|P_t\|^2$}, x tick label style={font=\small,text width=1cm,xshift=0.35cm}, enlarge y limits=0.1, y tick label style={font=\small}, ymin=0,ymax=1,xtick={0, 0.2, 0.4, 0.6, 0.8,1.0},legend pos=south east,legend cell align=left,legend style={draw=none,font=\scriptsize},axis background/.style={}]
        \addplot[blue,only marks,mark size=1.2pt] table[y=PgUnwatermarked, x=l2norm] {data/illustration/KGW_OPT_1_3b/all.dat};

    \end{axis}
\end{tikzpicture}%
\begin{tikzpicture}[]
    \begin{axis}[scale=1,width=0.32\linewidth,height=0.2\linewidth, xlabel style={yshift=0.5em},ymin=0,ymax=0.5,yticklabel= {\pgfmathprintnumber\tick}, ylabel={\small $S_t$},ylabel near ticks, enlarge x limits=0.05, xticklabel= {\pgfmathprintnumber\tick}, xlabel={\small $\|P_t\|^2$}, x tick label style={font=\small,text width=1cm,xshift=0.35cm}, enlarge y limits=0.1, y tick label style={font=\small}, xmin=0,xmax=1,xtick={0, 0.2, 0.4, 0.6, 0.8,1.0},legend pos=south east,legend cell align=left,legend style={draw=none,font=\scriptsize},axis background/.style={}]
       \addplot[blue,only marks,mark size=1.2pt] table[x=l2norm, y=Delta] {data/illustration/KGW_OPT_1_3b/all.dat};
   \end{axis}
\end{tikzpicture}

%% file: figure_scripts/fig_scatter_pg_delta_gumble_synthID.tex
\begin{tikzpicture}[]
    \begin{axis}[scale=1,width=0.3\linewidth,height=0.2\linewidth, title=Gumbel,ymin=-0.5,ymax=12,yticklabel= {\pgfmathprintnumber\tick}, ylabel={\small $S_t$},xlabel style={yshift=0.5em}, ylabel near ticks, enlarge x limits=0.05, title style={yshift=-0.5em}, xticklabel= {\pgfmathprintnumber\tick}, xlabel={\small $\|P_t\|^2$}, x tick label style={font=\small,text width=1cm,xshift=0.35cm}, enlarge y limits=0.1, y tick label style={font=\small}, xmin=0,xmax=1,xtick={0, 0.5, 1},legend pos=south east,legend cell align=left,legend style={draw=none,font=\scriptsize},axis background/.style={}]
       \addplot[blue,only marks,mark size=1.2pt] table[x=l2norm, y=Delta] {data/illustration/gumble_OPT_1_3b/all.dat};
   \end{axis}
\end{tikzpicture}
\begin{tikzpicture}[]
    \begin{axis}[scale=1,width=0.3\linewidth,height=0.2\linewidth, xlabel style={yshift=0.5em},ymin=0,title style={yshift=-0.5em}, ymax=0.4,yticklabel= {\pgfmathprintnumber\tick}, xlabel={\small $\|P_t\|^2$},ylabel near ticks, title={Tournament ($m=1$)},enlarge x limits=0.05, xticklabel= {\pgfmathprintnumber\tick}, ylabel={\small $S_t$}, x tick label style={font=\small,text width=1cm,xshift=0.35cm}, enlarge y limits=0.1, y tick label style={font=\small}, xmin=0,xmax=1,xtick={0,0.5, 1.0},legend pos=south east,legend cell align=left,legend style={draw=none,font=\scriptsize},axis background/.style={}]
        \addplot[blue,only marks,mark size=1.2pt] table[x=l2norm, y=Delta] {data/illustration/synthID_OPT_1_3b/all_m1.dat};
\end{axis}
\end{tikzpicture}
\begin{tikzpicture}[]
    \begin{axis}[scale=1,width=0.3\linewidth,height=0.2\linewidth, xlabel style={yshift=0.5em},title style={yshift=-0.5em}, ymin=0,ymax=0.4,yticklabel= {\pgfmathprintnumber\tick}, xlabel={\small $\|P_t\|^2$},ylabel near ticks, title={Tournament ($m=30$)},enlarge x limits=0.05, xticklabel= {\pgfmathprintnumber\tick}, ylabel={}, x tick label style={font=\small,text width=1cm,xshift=0.35cm}, enlarge y limits=0.1, y tick label style={font=\small}, xmin=0,xmax=1,xtick={0,0.5, 1.0},legend pos=south east,legend cell align=left,legend style={draw=none,font=\scriptsize},axis background/.style={}]
        \addplot[blue,only marks,mark size=1.2pt] table[x=l2norm, y=Delta] {data/illustration/synthID_OPT_1_3b/all_m30.dat};
\end{axis}
\end{tikzpicture}

%% file: figure_scripts/fig_scatter_l2_dtv.tex
\begin{tikzpicture}
    \begin{axis}[title=Green-Red, width=0.32\linewidth, height=0.2\linewidth, 
                 ymin=0, ymax=1.2, title style={yshift=-0.5em}, xlabel style={yshift=0.5em}, 
ylabel={\small $D_{TV}$}, ylabel near ticks, 
                 enlarge x limits=0.05, xtick={0, 0.2, 0.4, 0.6, 0.8, 1.0}, 
                 xlabel={\small $\|P_t\|^2$}, x tick label style={font=\small}, 
                 enlarge y limits=0.1, y tick label style={font=\small}, 
                 xmin=0, xmax=1, legend pos=north west, legend cell align=left, 
                 legend style={draw}]
        \addplot[blue, only marks, mark size=1.2pt] table[x=l2norm, y=tvd] {data/illustration/KGW_OPT_1_3b/all.dat};
        \addplot[red, only marks, mark size=1.2pt] table[=l2norm, y=tvdref] {data/illustration/KGW_OPT_1_3b/all.dat};
    \end{axis}
\end{tikzpicture}%
\begin{tikzpicture}
    \begin{axis}[title=Gumbel, width=0.32\linewidth, height=0.2\linewidth, title style={yshift=-0.5em}, xlabel style={yshift=0.5em}, 
                 ymin=0, ymax=1.2, ylabel={}, ylabel near ticks, 
                 enlarge x limits=0.05, xtick={0, 0.2, 0.4, 0.6, 0.8, 1.0}, 
                 xlabel={\small $\|P_t\|^2$}, x tick label style={font=\small}, 
                 enlarge y limits=0.1, y tick label style={font=\small}, 
                 xmin=0, xmax=1, legend pos=north west, legend cell align=left, 
                 legend style={draw}]
        \addplot[blue, only marks, mark size=1.2pt] table[x=l2norm, y=tvd] {data/illustration/gumble_OPT_1_3b/all.dat};
        \addplot[red, only marks, mark size=1.2pt] table[x=l2norm, y=tvdref] {data/illustration/gumble_OPT_1_3b/all.dat};

    \end{axis}
\end{tikzpicture}%
\begin{tikzpicture}
    \begin{axis}[title=Tournament, width=0.32\linewidth, height=0.2\linewidth, title style={yshift=-0.5em}, xlabel style={yshift=0.5em}, 
                 ymin=0, ymax=1.2, ylabel={}, ylabel near ticks, 
                 enlarge x limits=0.05, xtick={0, 0.2, 0.4, 0.6, 0.8, 1.0}, 
                 xlabel={\small $\|P_t\|^2$}, x tick label style={font=\small}, 
                 enlarge y limits=0.1, y tick label style={font=\small}, 
                 xmin=0, xmax=1, legend pos=north west, legend cell align=left, 
                 legend style={draw}]
        \addplot[blue, only marks, mark size=1.2pt] table[x=l2norm, y=tvd] {data/illustration/synthID_OPT_1_3b/all_m30.dat};
        \addplot[red, only marks, mark size=1.2pt] table[x=l2norm, y=tvdref] {data/illustration/synthID_OPT_1_3b/all_m30.dat};
    \end{axis}
\end{tikzpicture}%

%% file: figure_scripts/fig_scatter_st_dtv.tex
\begin{tikzpicture}
    \begin{axis}[width=0.32\linewidth, title style={yshift=-0.5em}, title=Green-Red,height=0.2\linewidth, 
                 ymin=0, ymax=1.0, xlabel style={yshift=0.5em},ylabel={\small $D_{TV}$}, ylabel near ticks,                  enlarge x limits=0.05, xtick={0, 0.2, 0.4}, 
                 xlabel={\small $S_t$}, x tick label style={font=\small}, 
                 enlarge y limits=0.1, y tick label style={font=\small}, 
                 xmin=0, xmax=0.5, legend pos=south east, legend cell align=left, 
                 legend style={draw=none}]
        \addplot[blue, only marks, mark size=1.2pt] table[x=Delta, y=tvd] {data/illustration/KGW_OPT_1_3b/all.dat};
    \end{axis}
\end{tikzpicture}%
\begin{tikzpicture}
    \begin{axis}[width=0.32\linewidth, title style={yshift=-0.5em}, xlabel style={yshift=0.5em},height=0.2\linewidth, title=Gumbel,
                 ymin=0, ymax=1.0, ylabel={}, ylabel near ticks, 
                 enlarge x limits=0.05, xtick={0, 5, 10}, 
                 xlabel={\small $S_t$}, x tick label style={font=\small}, 
                 enlarge y limits=0.1, y tick label style={font=\small}, 
                 xmin=-1, xmax=10, legend pos=south east, legend cell align=left, 
                 legend style={draw=none}]
        \addplot[blue, only marks, mark size=1.2pt] table[x=Delta, y=tvd] {data/illustration/gumble_OPT_1_3b/all.dat};
    \end{axis}
\end{tikzpicture}%
\begin{tikzpicture}
    \begin{axis}[width=0.32\linewidth, height=0.2\linewidth, xlabel style={yshift=0.5em},
                 ymin=0, ymax=1.0, title style={yshift=-0.5em}, title=Tournament ,ylabel={}, ylabel near ticks, 
                 enlarge x limits=0.05, xtick={0, 0.2, 0.4}, 
                 xlabel={\small $S_t$}, x tick label style={font=\small}, 
                 enlarge y limits=0.1, y tick label style={font=\small}, 
                 xmin=-0.05, xmax=0.4, legend pos=south east, legend cell align=left, 
                 legend style={draw=none}]
        \addplot[blue, only marks, mark size=1.2pt] table[x=Delta, y = tvd] {data/illustration/synthID_OPT_1_3b/all_m30.dat};
    \end{axis}
\end{tikzpicture}%
\\

%% file: sec/attack.tex
\section{Smoothing Attack}\label{sec:attack}
\label{sec:smoothing-attack}

Our objective is to design a watermark removal attack such that for any given input prompt, the algorithm returns some output text that \textbf{(i)} is of high-quality, i.e., comparable to those produced by the original un-watermarked model, while \textbf{(ii)} remaining free of watermarks, i.e., the generated text should evade watermark detection mechanisms. 

We design the \textit{Smoothing Attack}, based on our findings in Section~\ref{sec:limit}. Our attack proceeds as follows at each token position: \textbf{(i)} we first determine if the model is confident in its output token (it suffices to estimate $\|P_t\|^2$, according to our observation in Section~\ref{sec:detection-confidence}), and then \textbf{(ii)} we replace the token therein by the token sampled from a reference model (if the watermark is distortionary) or by the token re-sampled from the watermarked model (if the watermark is distortion-free). 

\paragraph{Model Access.}
We consider a practical scenario where the adversary has limited access to the targeted watermarked model and does not have access to the original un-watermarked model. Hence, he cannot estimate $\|P_t\|^2$ directly. We assume that at each token position, the adversary knows the top-\(K\) tokens that are most likely to be sampled and the corresponding probabilities (with \(K\) being a small constant, e.g., $1,5,10$) via the target watermarked model's API. This level of information is commonly accessible, even for closed-source models, e.g., OpenAI's API provides information on the most likely tokens and the corresponding probabilities.

\paragraph{Estimation of model's confidence.} For distortion-free watermarks (e.g., Gumbel and Tournament sampling), we can directly observe the unchanged top-\(K\) probabilities of the un-watermarked model. We then estimate $\|P_t\|^2$ from the probabilities associated with the top-\(K\) most probable tokens in $\mathcal{V}_\text{Top-K}$, expressed as $\|P_t\|^2 \approx
\sum_{v \in \mathcal{V}_{\text{Top-K}}} P(v)^2 
\;+\;
\frac{1}{|\mathcal{V}| - K}\cdot \left(1 - \sum_{v \in \mathcal{V}_{\text{Top-K}}} P(v)\right)^2,$ where we have assumed that the residual probability mass \(1 - \sum_{v \in \mathcal{V}_{\text{Top-K}}} P(v)\) is uniformly distributed across the remaining \( |\mathcal{V}| - K \) tokens whose probabilities are unobserved.

For distortionary watermarks (e.g., the Green-red list), we estimate $\|P_t\|^2$ from the top-\(K\) probabilities observed in the modified $\widetilde P_t$. In such cases, we first compute an estimation for the watermarked model's  $\|\widetilde P_t\|^2\approx \sum_{v \in \mathcal{V}_{\text{Top-K}}} P(v)^2 +
\frac{1}{|\mathcal{V}| - K}\cdot \left(1 - \sum_{v \in \mathcal{V}_{\text{Top-K}}} \widetilde P(v)\right)^2$. Again, we have assumed that the residual probability mass $1 - \sum_{v \in \mathcal{V}_{\text{Top-K}}} \widetilde P(v)$ is evenly distributed to the unobserved tokens. Next, we transform this estimation to the original unwatermarked model by computing $\|P_t\|^2 \approx \beta\cdot \|\widetilde P_t\|^2$ with $
\beta={\frac{\left((1 - \gamma) + \gamma e^\delta\right)^2}{(1 - \gamma) + \gamma e^{2\delta}}}$. Detailed derivations are deferred to Appendix~\ref{appendix:estimate_l2} and~\ref{appendix:estimate_l2_using_topk}.

\paragraph{Normalization of confidence.} We convert the above estimated $\|P_t\|^2$ into a confidence score in $[0,1]$, denoted as $c$. To establish the upper and lower bounds for normalization, denoted as $U$ and $L$, we obtain $N$ (e.g., $N=200$) random samples for $\|P_t\|^2$, computed on the watermarked model with random prefixes. We then set $U$ and $L$ as the largest and smallest values among the samples, respectively. Next, we construct $100$ bins from the range $[L,U]$ and then map the estimated $\|P_t\|^2$ into the $i$-th bin with $i\in [1,100]$. The confidence score \(c\) is then computed as \(
c = \frac{i}{100}.
\) Based on $c$, we decide whether to adopt the token output by the watermarked model at this token position. 

\paragraph{Smoothing the watermark.} When the watermarking is distortion-free, with probability $c^{\alpha}$, we adopt this token; with probability $(1-c^{\alpha})$, we randomly choose another token by re-sampling the token based on the observed top$-K$ probabilities of the target watermarked model, and then put it into the token position. 

When the watermarking is distortionary, we first query a reference model (in our experiments, the reference model is much smaller than the watermarked model) using the same prefix and then obtain the top-$K$ probabilities from the reference model. After that, we compute a weighted probability distribution from the top-$K$ probabilities of the watermarked model and the reference model. For the reference model, the weight is assigned to $1-c^{\alpha}$; for the watermarked model, the weight is assigned to $c^{\alpha}$. We then sample from the weighted distribution and adopt the output token. 

Here we have used a constant exponential factor $\alpha > 0$ to control the smoothness-level of our attack. In particular, when $\alpha$ is large, our attack inclines to adopt the token from the watermarked model. On the other hand, when $\alpha$ is small, our attack inclines to replace the output token. 

\paragraph{Applicability.}
Our attack does not require detailed knowledge of the watermarking strategy. Instead, it either samples a new token or adopts the output token from the watermarked model, based on the confidence score $c$. In addition, it is computationally efficient, and the main overhead incurred when estimating the upper and lower bounds for $c$, i.e., $U$ and $L$. This can be done by querying the watermarked model a few hundred times.

%% file: sec/exp.tex
\input{data/tab_avg_combined.tex}

\section{Experiments}\label{sec:exp}

\noindent\textbf{Models and datasets.}
We attack the open-sourced models including, Llama3.1-8B~\cite{dubey2024llama}, the OPT model family~\cite{zhang2022opt} (from 1.3B to 30B parameters), and Qwen2-1.5B~\cite{chu2024qwen2}. When attacking distortionary watermarking algorithms on Llama3, OPT models, and Qwen2, we use Llama3-1B~\cite{dubey2024llama}, OPT-125M~\cite{zhang2022opt}, and Qwen2-0.5B~\cite{chu2024qwen2}  as the reference models, respectively. Following prior work~\cite{kirchenbauer23a,pan2024markllm}, we evaluate on the C4 dataset~\cite{raffel2020exploring}. For each text, its first $30$ tokens of texts serve as the prompt, and the task is to generate the subsequent $200$ tokens. The results are averaged over $100$ prompts. Our experiments are run on RTX-Titan GPUs.

\paragraph{Watermarking algorithms.}
We evaluate against $10$ representative watermarking algorithms, covering distortionary and distortion-free watermarks, including  KGW~\cite{kirchenbauer23a}, Unigram~\cite{zhao2023provable}, UPV~\cite{liu2023unforgeable}, X-SIR~\cite{hewatermarks2024}, DIP~\cite{wu2023dipmark}, SWEET~\cite{lee2023wrote},
 EWD~\cite{lu2024entropy} Unbiased~\cite{huunbiased},  SynthID~\cite{dathathri2024scalable} (which leverages Tournament sampling), and Gumbel~\cite{aaronson2023watermarking}. The implementations are based on the MarkLLM toolkit~\cite{pan2024markllm}. For Gumbel and X-SIR, we evaluate it on OPT models (Gumbel requires more than 100 GB of GPU memory when running on Qwen and Llama due to their large vocabulary sizes and X-SIR's official code does not support Llama and Qwen models for now).

\paragraph{Attack baseline.} 
We include the strongest watermark removal attack, \textit{Paraphrasing Attack} that uses GPT-3.5-turbo to paraphrase the watermarked text~\cite{piet2023mark},  as a competitor.
As a comparison, our \textit{Smoothing Attack} leverages only much smaller reference models (when attacking distortionary watermarks).

\paragraph{Metrics.} 
We evaluate the performance of attacks in terms of watermark removal and text quality preservation. For watermark removal, we report the true positive rate of watermark detection, i.e., TPR, (lower means the attack is better) when the false positive rate, i.e., FPR, is less than 1\%. In this case, TPR is $1\%$ for un-watermarked models and $100\%$ without attacks. For text quality, we follow prior work~\cite{kirchenbauer23a,pan2024markllm} to measure the perplexity (lower means better text quality). 

For our attack, we set $\alpha$ to $1.0$ and use the top-$10$ most likely tokens and their probabilities from the watermarked model and reference model by default, unless specified otherwise. We also report the diversity of the generated text, following~\cite{kirchenbauer2023reliability} (higher means better). We defer detailed descriptions to the appendix.

\paragraph{Performance in watermark removal.} We present the main results in Tables~\ref{tab:combined_baselines} and~\ref{tab:effectiveness_small_xsir_gumbel}. Our \textit{Smoothing Attack} successfully removes watermarks for most of the cases (achieving a low TPR around $5\%$ and even $0$ sometimes) across all watermarked models and watermarking algorithms. 

Our attack also outperforms the strong paraphrasing attack in terms of TPR. Notably, for the OPT-1.3B model with the Unigram watermark (see Table~\ref{tab:combined_baselines}), the detector can successfully detect a $53\%$ fraction of the watermarked text even after paraphrasing, while it only identifies a $5\%$ fraction if using our attack. We note that our attack achieves this by using a much smaller reference model, OPT-125M.

\begin{table}[t]
\centering
\caption{Performance of watermark removal attacks on OPT-1.3B with Gumbel~\cite{aaronson2023watermarking} and X-SIR~\cite{hewatermarks2024} watermarks (with FPR less than 1\%).}
\label{tab:effectiveness_small_xsir_gumbel}
\begin{tabular}{lcccc}
\toprule
Watermark & Attack & TPR (in \%) & PPL & Div \\\midrule
\multirow{3}{*}{Gumbel} & None & 98 & 2.96 & 4.35 \\
 & Paraphrasing & 13 & 14.21 & 11.13 \\
 & Smoothing & 9 & 19.25 & 8.30 \\\hline
\multirow{3}{*}{X-SIR} & None & 94 & 13.99 & 7.96 \\
 & Paraphrasing & 34 & 14.13 & 8.80 \\
 & Smoothing & 9 & 9.47 & 6.75  \\\bottomrule
\end{tabular}
\end{table}

\paragraph{Performance in text quality.}
Our attack also preserves the text quality, meaning that it achieves a low perplexity without decreasing the diversity too much. For example, when attacking the Unigram watermark on OPT-1.3B (see Table~\ref{tab:combined_baselines}), our attack achieves a perplexity of $9.44$ (much better than $14.51$ of paraphrasing attack) and a diversity of $6.73$ (only slightly worse than $8.75$ of paraphrasing attack) while achieving a TPR of $5\%$ (much better than $50\%$ of paraphrasing attack). More detailed boxplots for the text quality metrics are in \autoref{fig:ppl_opt} and~\autoref{fig:diversity_opt} in Appendix~\ref{appendix:text_quality_evaluation}.

For Gumbel sampling (see Table~\ref{tab:effectiveness_small_xsir_gumbel}), our attack increases the perplexity. The main reason is that Gumbel sampling tends to output repeated content, which sometimes leads to better perplexity, but lower diversity. We give concrete output text samples in Table~\ref{tab:gumbel_example} in Appendix~\ref{appendix:text_example}, to show that our attack generates better texts than the watermarked model. 

\begin{table}[t!]
\centering
\begin{minipage}[t]{0.95\linewidth}
\centering
\caption{Impact of $K$ and $\alpha$ on the performance of \textit{Smoothing Attack} on OPT-1.3B with Unigram watermark (with FPR less than 1\%).}
\label{tab:impact_of_K_alpha_opt_unigram}
\begin{tabular}{lcccc}
\toprule
$K$ & $\alpha$  & TPR (in $\%$) & PPL & Div \\
\midrule
\multirow{4}{*}{Fixed to 10} & 0.5 & 42 & 9.9  & 6.86 \\
&1.0 & 5  & 9.44 & 6.73 \\
 &2.0 & 0  & 9.38 & 6.58 \\
 &3.0 & 1  & 9.25 & 6.43 \\
 \midrule\midrule
1  & \multirow{4}{*}{Fixed to 1} & 18 & 3.21  & 4.62 \\
5  & & 10 & 7.46  & 6.11 \\
10 & & 5  & 9.44  & 6.73 \\
15 & & 5 & 11.73 & 7.11 \\
\bottomrule
\end{tabular}
\end{minipage}
\end{table}

The diversity of the text generated by our attack is slightly worse than those generated from the paraphrasing attack (and also sometimes worse than the watermarked model). The main reason is that our smoothing attack selects the next token only from Top-$K$ most likely tokens, while the paraphrasing attack and the watermarked model sample the token from the whole vocabulary (which, in turn, may increase the unpredictability and degrade the text quality). Increasing $K$ can resolve this issue, e.g., see Table~\ref{tab:impact_of_K_alpha_opt_unigram} for the results on the Unigram watermark (more in appendix). The trade-off is that larger $K$'s require more model access.  

Our smoothing attack also achieves a lower (hence, better) PPL compared to the unwatermarked text. This is because, for the unwatermarked text, the token is sampled from the whole vocabulary, which makes the output unpredictable sometimes and even erratic. In our smoothing attack, the text is selected from the top-$K$ most likely tokens from the target watermarked model and reference model, preventing extremely unlikely tokens from being output. 

\paragraph{Ablation studies. }
The overall observation is that increasing $K$ leads to texts with worse perplexity, but better diversity and better TPR. We note that even with $K=1$, our attack is still effective. In Table~\ref{tab:impact_of_K_alpha_opt_unigram}, the TPR is only $18\%$, lower than that of GPT-3.5-turbo, which is $53\%$. Increasing $\alpha$ makes our attack tend to forbid the uncertain tokens in the watermarked text when the model is not confident. Hence, increasing $\alpha$ in general improves the perplexity while reducing the diversity sometimes (e.g., see~\autoref{tab:impact_of_K_alpha_opt_unigram}). At the same time, the detection rate also decreases, since the watermark traces are more likely to get removed. (Note that this event happens at probability $1-c^{\alpha}$, which increases as $\alpha$ increases for $c\in (0,1]$). By adjusting $\alpha$, the adversary can balance between watermark removal and text diversity. More details are deferred to Appendix~\ref{appendix:impact_alpha_k}.

We also conduct ablation studies on the model size within the OPT model family (from $1.3$B to $30$B parameters). For all target models, we use the same reference model OPT-125M, which is much smaller than the target model. Our finding is that the model size has negligible influence to the performance of our attack. For instance, the TPR of our attack against X-SIR increases from $13\%$ to $16\%$ when the size of the target model grows from $1.3$B to $30$B. We refer to Appendix~\ref{appendix:impact_of_model_size} for more details. 

\paragraph{Summary.} Overall, our \textit{Smoothing Attack} excels at removing the watermark (measured as TPR in watermark detection) while preserving the quality of the generated text (measured as perplexity and diversity), demonstrating universal applicability across different models and watermarking algorithms. Parameters $K$ and $\alpha$ control the trade-off between watermark removal and text quality preservation. We discuss possible defenses to our attack in Appendix~\ref{appendix:possible_defense}.

%% file: data/tab_avg_combined.tex
\begin{table*}[ht!]
\centering
\caption{Performance of watermark removal attacks on OPT-1.3B, Llama3-8B, and Qwen-1.5B. The false positive rate on the unwatermarked text is less than 1\%. We show the true positive rate in $\%$ (TPR), perplexity (PPL), and diversity (Div) for each watermarking algorithm for different models and against different attacks. }
\label{tab:combined_baselines}
\begin{tabular}{llcccccccccc}
\toprule
\small
\multirow{2}{*}{Watermark} & \multirow{2}{*}{Attack} & \multicolumn{3}{c}{OPT-1.3B} & \multicolumn{3}{c}{Llama3-8B} & \multicolumn{3}{c}{Qwen2-1.5B} \\
\cmidrule(lr){3-5} \cmidrule(lr){6-8} \cmidrule(lr){9-11}
 & & TPR & PPL & Div & TPR & PPL  & Div & TPR  & PPL & Div \\\midrule
Un-watermarked & - & 1 & 11.39 & 8.22 & 1 & 3.47 & 6.82 & 1 & 12.26 & 8.10 \\
Reference & - & 1 & 19.57 & 7.69 & 1 & 4.40 & 6.52 & 1 & 16.02 & 8.06 \\\hline
\multirow{3}{*}{KGW~\cite{kirchenbauer23a}} &  None & 100 & 14.61 & 8.07 & 99 & 4.60 & 6.92 & 100 & 16.46 & 8.11 \\
 & Paraphrasing & 3 & 14.82 & 9.56 & 2 & 5.35 & 8.0 & 2&10.45&9.42\\
 & Smoothing & 0 & 9.57 & 6.72 & 2 & 3.20 & 5.63 & 0 & 8.02 & 6.91 \\\hline
\multirow{3}{*}{Unigram~\cite{zhao2023provable}} &  None & 100 & 14.99 & 7.29 & 99 & 4.61 & 6.56 & 100 & 15.41 & 7.37 \\
 & Paraphrasing & 53 & 14.51 & 8.75 & 54 & 5.60 & 8.02 & 5 & 10.40 & 8.56 \\
 & Smoothing & 5 & 9.44 & 6.73 & 24 & 3.10 & 5.44 & 1 & 7.77 & 6.71 \\\hline
\multirow{3}{*}{SynthID~\cite{dathathri2024scalable}} &  None & 100 & 7.12 & 7.41 & 99 & 4.83 & 7.31 & 100 & 6.94 & 7.05 \\
 & Paraphrasing & 1 & 10.57 & 9.11 & 1 & 5.62 & 8.18 & 1 & 6.90 & 8.43 \\
  & Smoothing & 0 & 10.40 & 8.64 & 0 & 3.40 & 6.86 & 0 & 10.21 & 8.04 \\\hline

\multirow{3}{*}{DIP~\cite{wu2023dipmark}} &  None & 100 & 13.73 & 8.44 & 84 & 4.03 & 7.35 & 100 & 14.34 & 8.27 \\
 & Paraphrasing & 0 & 13.95 & 9.25 & 0 & 5.25 & 8.34 & 2 & 10.10 & 8.85 \\
 & Smoothing & 6 & 9.34 & 6.84 & 6 & 3.17 & 5.67 & 11 & 7.62 & 6.92 \\\hline
\multirow{3}{*}{Unbiased~\cite{huunbiased}} &  None & 100 & 13.61 & 8.29 & 84 & 4.02 & 7.29 & 100 & 14.64 & 8.21 \\
 & Paraphrasing & 3 & 14.45 & 10.39 & 2 & 5.36 & 8.57 & 1 & 9.97 & 8.82 \\
 & Smoothing & 27 & 9.19 & 6.84 & 5 & 3.17 & 5.75 & 5 & 7.68 & 6.94 \\\hline
\multirow{3}{*}{UPV~\cite{liu2023unforgeable}} &  None & 99 & 11.65 & 8.22 & 83  & 4.38 & 6.80 & 86  & 11.93 & 7.49 \\
 & Paraphrasing & 34  & 13.73 & 9.92 & 2  & 5.43 & 8.00 & 2  & 9.03 & 8.58 \\
 & Smoothing & 20  & 10.01 & 6.89 & 1  & 3.12 & 5.49 & 0  & 8.16 & 6.91 \\\hline
\multirow{3}{*}{EWD~\cite{lu2024entropy}} &  None & 100  & 15.23 & 7.92 & 100  & 4.56 & 6.71 & 100  & 16.31 & 7.85 \\
 & Paraphrasing & 0  & 14.95 & 9.95 & 7  & 5.73 & 7.83 & 1  & 10.18 & 9.28 \\
 & Smoothing & 0  & 9.93 & 6.78 & 3  & 3.13 & 5.38 & 0  & 7.82 & 6.85 \\\hline
\multirow{3}{*}{SWEET~\cite{lee2023wrote}} &  None & 100  & 14.36 & 8.02 & 99  & 4.53 & 6.69 & 100  & 15.89 & 7.65 \\
 & Paraphrasing & 0  & 14.57 & 9.45 & 14  & 5.64 & 8.05 & 4  & 10.18 & 9.30 \\
 & Smoothing & 0  & 9.59 & 6.72 & 4  & 3.09 & 5.40 & 0  & 7.85 & 6.92 \\
 \bottomrule
\end{tabular}
\end{table*}

%% file: sec/conclusion.tex
\section{Conclusion}
We revealed limitations in existing watermarks for language models and examined their robustness against watermark removal attacks. We introduced \textit{Smoothing Attack}, a novel method that leverages model confidence to selectively remove watermark traces while preserving text quality. Comprehensive evaluations demonstrated that \textit{Smoothing Attack} can completely remove watermarks, outperforming the state-of-the-art attack and highlighting a critical gap in current watermarks, and calling for more robust solutions.

%% file: data/tab_notation.tex
\begin{table}[ht]
    \centering
    \caption{Table of notation definitions and their locations.}
    \label{tab:notation}
    \renewcommand{\arraystretch}{1.3}
    \resizebox{0.9\linewidth}{!}{
    \begin{tabular}{cp{30em}c}
        \hline
        \textbf{Notation} & \textbf{Meaning} & \textbf{Definition Location} \\
        \hline
        $M$ & Auto-regressive language model (LM), which generates text sequentially based on a given prompt. & Section~\ref{sec:background} \\
        $\widetilde{M}$ & Watermarked model, a variant of $M$ that embeds watermarks into generated text. & Section~\ref{sec:detection-confidence} \\
        $\mathcal{V}$ & Vocabulary of the LM, the set of all possible tokens that can be generated. & Section~\ref{sec:background} \\
        $t$ & Token position in the generated sequence, indicating the index of a specific token. & Section~\ref{sec:background} \\
        $l_t(v)$ & Logit assigned by the model to token $v$ at position $t$ before applying softmax. & Eq.~\eqref{eq:soft-max} \\
        $P_t(v)$ & Probability of token $v$ at position $t$ after applying the softmax function. & Eq.~\eqref{eq:soft-max} \\
        $\widetilde{P}_t(v)$ & Modified probability distribution in the watermarked model after logit manipulation. & Eq.~\eqref{eq:modified_p_green_list} \\
        $(v_1, \dots, v_T)$ & Sequence of tokens forming the output text from the language model. & Section~\ref{sec:background} \\
        $d(v_1, \dots, v_T)$ & Detection score function used to determine whether a text is watermarked. & Section~\ref{sec:background} \\
        $\tau$ & Threshold value for watermark detection; if $d(v_1, \dots, v_T) > \tau$, the text is classified as watermarked. & Section~\ref{sec:background} \\
        $\mathcal{G}_t$ & Green list, a subset of vocabulary containing tokens whose logits are increased in green-red list watermarking. & Section~\ref{sec:background} \\
        $\gamma$ & Fraction of the vocabulary included in the green list $\mathcal{G}_t$, determining the probability of token selection. & Section~\ref{sec:background} \\
        $\delta$ & Logit increase applied to tokens in the green list $\mathcal{G}_t$, influencing token selection probabilities. & Eq.~\eqref{eq:modified_p_green_list} \\
        $T$ & Length of the generated sequence, i.e., the total number of tokens in the output text. & Section~\ref{sec:background} \\
        $u_t(v)$ & Randomly sampled value from $[0,1]$ for token $v$ in Gumbel sampling watermarking. & Section~\ref{sec:background} \\
        $v_t^*$ & Token selected using Gumbel sampling watermarking by maximizing a transformed probability. & Section~\ref{sec:background} \\
        $S_t$ & Contribution of the token at position $t$ to the overall watermark detection score. & Eq.~\eqref{eq:contribution-green} \\
        $\mathbb{E}_{v \sim P_t}[\mathbf{1}\{v \in \mathcal{G}_t\}]$ & Expected probability mass assigned to green tokens at position $t$ from probability distribution $P_t$. & Eq.~\eqref{eq:contribution-green} \\
        $\|P_t\|^2$ & $\mathcal{L}_2$ norm of the probability vector, measuring model confidence at position $t$. A higher value means greater confidence. & Section~\ref{sec:detection-confidence} \\
        $D_{TV}(P_t, \widetilde{P}_t)$ & Total variation distance between original and watermarked probability distributions, measuring distortion. & Section~\ref{sec:detection-confidence} \\
        $D_{TV}(P_t, P^{\text{ref}}_t)$ & Total variation distance between the original model and a reference model’s probability distributions. & Section~\ref{sec:detection-confidence} \\
        $K$ & Number of most probable tokens that the adversary has access to from the watermarked model. & Section~\ref{sec:smoothing-attack} \\
        $\mathcal{V}_{\text{Top-K}}$ & Set of top-$K$ most probable tokens observed by the adversary. & Section~\ref{sec:smoothing-attack} \\
        $\beta$ & Scaling factor used to estimate $\|P_t\|^2$ from watermarked probabilities in Green-red list watermarking. & Section~\ref{sec:smoothing-attack} \\
        $c$ & Normalized confidence score in $[0,1]$ based on estimated $\mathcal{L}_2$ norm. & Section~\ref{sec:smoothing-attack} \\
        $U, L$ & Upper and lower bounds for normalizing $\mathcal{L}_2$ norms into the confidence score $c$. & Section~\ref{sec:smoothing-attack} \\
        $\alpha$ & Exponential factor controlling the aggressiveness of the smoothing attack. A larger $\alpha$ favors keeping watermarked tokens, while a smaller $\alpha$ favors replacement. & Section~\ref{sec:smoothing-attack} \\
        $P^{\text{ref}}_t$ & Token probability distribution from a much smaller, un-watermarked reference model. & Section~\ref{sec:smoothing-attack} \\
        \hline
    \end{tabular}}
\end{table}

%% file: sec/appendix_additional_results.tex
\section{More on Related Work}
\label{appendix:related}

\paragraph{Variations of Green-red list watermark.} Different variations of Green-red list watermark, e.g., see~\cite{kirchenbauer2023reliability,lee2023wrote,liu2023unforgeable,wu2023dipmark,huotoken,zhou2024bileve,lu2024entropy,liu2024a,hewatermarks2024,zhao2023provable,kirchenbauer23a}, mainly differ in the assignment of the green lists and the detection process. In particular, the assignment of $\mathcal{G}_t$ could depend on the prefix, e.g., the preceding $h$ tokens in the generated text. When $h=0$, we say the assignment is context-independent and is referred to as the \textit{Unigram} watermark~\cite{zhao2023provable}; when $h=1$, the assignment depends on the previous token and is referred to as the \textit{KGW} watermark~\cite{kirchenbauer23a}

\paragraph{Scalable Tournament sampling.} As shown in their paper, the original tournament process in~\cite{dathathri2024scalable} can be costly to implement, as there are $O(2^m)$ times of sampling and pair-wise comparison of tokens. Instead, they obtain a modified distribution for tokens. With $\widetilde{P}^{(0)}_t=P_t$, they iteratively compute
$\widetilde{P}^{(l)}_t(v) = \Big(1+g^{(l)}(v, r_t) 
-\sum_{v'\in\mathcal{V}} \big(g^{(l)}(v', r_t)\cdot P^{(l-1)}_t(v') \big)\Big)\cdot \widetilde{P}^{(l-1)}_t(v),$ for $l=1,\ldots,m$, and then sample the token from $\widetilde{P}^{(m)}$. 

\paragraph{Distortion-free watermark.} There are also other distortion-free watermarks, which aim to preserve the original model's token distribution and avoid detectable shifts in probabilities of output tokens, e.g., see \cite{huunbiased,zhao2024permute,fu2024gumbelsoft,christ2023undetectable,fairoze2023publicly,christ2024pseudorandom,cohen2024watermarking,ghentiyala2024new,golowichedit,dathathri2024scalable,wu2023dipmark}. 

\paragraph{Comparison with paraphrasing attacks.} When attacking OPT models (from 1.3B to 30B parameters), our attack only leverages the OPT-125M as the reference model when attacking distortionary watermarks such as the Unigram watermark. When attacking distortion-free watermarks, our attack sometimes resamples from the target watermarked model. In either case, the resource used in our attack is significantly smaller than the state-of-the-art paraphrasing attack, which uses the much larger GPT-3.5-turbo. Despite using fewer resources, our approach achieves higher watermark removal rates and comparable text quality. This highlights that even resource-limited adversaries can thwart watermarks, underscoring the need for stronger watermark defenses.

\section{More on Experiments}
\label{appendix:exp}
\subsection{Implementation}\label{appendix:watermark_alg}
We evaluate the smoothing attack on eight different watermarking algorithms, including KGW~\cite{kirchenbauer23a}, Unigram~\cite{zhao2023provable}, SWEET~\cite{lee2023wrote}, UPV~\cite{liu2023unforgeable}, EWD~\cite{lu2024entropy}, X-SIR~\cite{hewatermarks2024},  DIP~\cite{wu2023dipmark}, Unbiased~\cite{huunbiased}, SynthID~\cite{dathathri2024scalable} and Gumbel~\cite{aaronson2023watermarking}. We use the implementations and default configurations provided by MarkLLM~\cite{pan2024markllm}. For completeness, we provide details of the algorithms below.

\begin{itemize}
\item  KGW~\cite{kirchenbauer23a}: The green set $\mathcal{G}_t$ at each position $t$ is selected based on the previous $h$ tokens and a secret key known to the service provider. The hyperparameters are set as follows: $\gamma = 0.5$, $\delta = 2.0$, and $h = 1$.

\item Unigram~\cite{kirchenbauer23a}: The green set $\mathcal{G}_t$ is fixed for each token $t$ and each prefix, depending solely on the secret key known to the service provider. No dynamic updates are performed based on previous tokens. The parameters are: $\gamma = 0.5$, $\delta = 2.0$.

\item  SWEET~\cite{lee2023wrote}: A shift is applied only when the entropy of the probability distribution at position $t$ is high, improving text quality, particularly for code generation tasks. The parameters are set as: $\gamma = 0.5$, $\delta = 2.0$, the entropy threshold is 0.9, and $h = 1.0$.

\item UPV~\cite{liu2023unforgeable}: The green token selection process is similar to the previous approaches. However, this method requires training two additional models: a generator network to separate red and green tokens and a detector network for classification based on the input text. The watermarks are introduced using $\gamma = 0.5$, $\delta = 2.0$, and $h = 1.0$. The detector produces a binary prediction rather than a continuous score like a z-score.

\item EWD~\cite{lu2024entropy}: Watermark introduction follows a similar process as the previous methods. The hyperparameters are $\gamma = 0.5$, $\delta = 2.0$, and $h = 1.0$. During detection, tokens are assigned different weights based on their entropy, with higher entropy tokens receiving greater weight to improve detectability in low-entropy scenarios.

\item  X-SIR~\cite{hewatermarks2024}: Instead of operating at the token level, the red-green partition is applied at the level of semantic clusters, grouping similar words together and adding bias at the group level. This improves robustness against Cross-lingual Watermark Removal Attacks (CWRA). 

\item DIP~\cite{wu2023dipmark}: Similar to Kirchenbauer et al. (2023), this method selects green tokens but uses a distribution-preserving reweight function to adjust token probabilities. This increases the probability of green tokens while maintaining the overall distribution. The reweighting is controlled by the parameter $\alpha$. The hyperparameters are set as $\gamma = 0.5$, $h = 5$. 
\end{itemize}

\paragraph{Implementation of the paraphrasing attack.}
We include the strongest baseline that paraphrases the given text based on the GPT-3.5-turbo~\cite{piet2023mark}, denoted as P-GPT3.5 using the prompt: ``Please rewrite the following text:''. As shown in~\cite{kirchenbauer2023reliability}, GPT-3.5-turbo is more powerful in removing the watermarks compared to Dipper model~\cite{krishna2023paraphrasing}.

\paragraph{Text quality metric.}
We use Llama3-8B, Qwen2-7B, and OPT-2.7B to evaluate the perplexity of the text generated from Llama3, Qwen2, and OPT models. We also report the log diversity of the text~\cite{welleckneural,kirchenbauer2023reliability,li2022contrastive}, following the definition in~\cite{kirchenbauer2023reliability} considering the 2-gram, 3-gram, and 4-gram repetition in the generated text. A higher diversity score represents a more diverse text. 

\subsection{Performance of the smoothing attack}
\label{appendix:effectiveness_smoothing_attack}
\autoref{fig:all_baseline_trade-off} shows three scatterplots of TPR vs.\ PPL for text generated under different watermarking and attack settings. Each point is colored by the watermarking method and corresponds to one of three models (OPT-1.3B, Llama3-8B, Qwen2-1.5B). Overall, the smoothing attack yields substantially lower TPR relative to the watermarked setting, demonstrating its performance at watermark removal. Notably, smoothing’s TPR is on par with that of the paraphrasing attack, which uses a more powerful model (GPT-3.5-turbo). In terms of perplexity (PPL), smoothing also generates text that is competitive with (and sometimes lower than) both the watermarked text and the paraphrased text, indicating that it preserves text quality while removing the watermark.

\begin{figure}
\centering
\input{figure_scripts/fig_scatter_trade-off.tex}
\caption{Each subfigure shows how the true positive rate (TPR) varies with perplexity (PPL) for a specific attack. No attack (a) corresponds to watermarked text without modifications, paraphrasing (b) uses GPT-3.5-turbo to rewrite the text, and smoothing (c) randomly replaces some tokens to remove the watermark. Colors indicate the particular watermarking method and each point corresponds to one of three models (OPT-1.3B, Llama3-8B, Qwen2-1.5B).}
\label{fig:all_baseline_trade-off}
\end{figure}

\begin{figure*}[h!]
\centering
\begin{minipage}{\textwidth}
\centering
\input{figure_scripts/fig_boxplot_ppl.tex}
\vspace{-12pt}
\caption{Text Quality Comparison – Perplexity (OPT-1.3B). Box plots of perplexity for text generated from different sources, with perplexity computed using the OPT-2.7B model. Our smoothing attack produces text with quality comparable to, and in some cases better than, that of the watermarked model.}
\label{fig:ppl_opt}
\end{minipage}
\begin{minipage}{\textwidth}
\centering
\input{figure_scripts/fig_boxplot_diversity.tex}
\vspace{-12pt}
\caption{Text Quality Comparison – Diversity (OPT-1.3B). Box plots of text diversity for outputs generated from different sources. Our smoothing attack produces text with diversity comparable to, and in some cases lower than, that of the watermarked model due to its constrained selection process.}
\label{fig:diversity_opt}
\end{minipage}
\end{figure*}

\subsection{Text quality evaluation}
\label{appendix:text_quality_evaluation}
\autoref{fig:ppl_opt} and \autoref{fig:diversity_opt} present boxplots of the perplexity (PPL) and diversity of text generated from different sources using the OPT-1.3B model. We observe that the smoothing attack generally yields text with lower PPL than the watermarked model, except in cases involving the Gumbel watermark. This suggests that, according to the PPL metric, the smoothing attack can generate high-quality text. In terms of diversity, the constrained selection process—where sampling is restricted to the top-K candidates from both the reference and target models—results in lower diversity for the smoothing attack. These findings are consistent with the average PPL results reported in Table~\ref{tab:combined_baselines} in the main paper.

In addition, we compute the P-SP score~\cite{wieting2022paraphrastic}, which quantifies the similarity between pairs of texts in the embedding space, with higher scores indicating greater similarity. Specifically, we calculate P-SP scores for text generated from different sources and visualize the results in the heatmap shown in~\autoref{fig:heatmap_psp}. We observe that, aside from the paraphrasing case, texts from different sources generally exhibit low similarity. For instance, text generated by the watermarked model has a P-SP score of 53.6 on Unigram, whereas the similarity between the watermarked text and its paraphrased version reaches 82.3. Our smoothing attack produces a P-SP score (measuring similarity between text from the smoothing attack and unwatermarked text) comparable to that of the watermarked text (measuring similarity between watermarked text and unwatermarked text). The generally low P-SP scores across different sources reflect the natural variability in generated responses, as multiple reasonable outputs can exist for the same prompt. Therefore, P-SP metrics may not be a reliable measure for assessing text quality degradation due to watermarking or smoothing.

\begin{figure*}[h!]
\begin{center}
\input{figure_scripts/fig_heatmap_psp.tex}
\caption{Text Quality Comparison – P-SP (OPT-1.3B). Heatmap comparing the similarity of text generated by different models in the sentence embedding space. Text from the watermarked model has a low similarity score compared to unwatermarked text, reflecting the inherent variability in generated responses. However, the paraphrased text (Paraphrasing vs. watermarked) exhibits a high similarity score, suggesting that the P-SP metric is more suitable for evaluating paraphrasing rather than assessing text quality degradation due to watermarking or smoothing.}
\label{fig:heatmap_psp}
\end{center}
\end{figure*}

\input{data/tab_impact_k_opt_rotated.tex}
\input{data/tab_impact_alpha_opt_rotated.tex}

\input{data/tab_impact_k_llama_rotated.tex}

\input{data/tab_impact_alpha_llama_rotated.tex}
\subsection{Effect of $K$ and $\alpha$}
\label{appendix:impact_alpha_k}
\autoref{tab:impact_of_k_opt} and \autoref{tab:impact_of_k_llama} show the performance of smoothing attacks against different watermarking algorithms under varying values of \( K \). In a smoothing attack, the adversary has access only to the top-\( K \) tokens and their probabilities from both the reference and target models. Even with \( K=1 \), the attack can drastically reduce the true positive rate (TPR) from 99\% (without any attack) to an extremely low value, sometimes reaching 0\%. This indicates that even with minimal access to both models, the smoothing attack can effectively remove watermarks. Furthermore, we observe that increasing \( K \) leads to more diverse text generation, as discussed in the main paper. This is because a higher \( K \) provides the attack with a larger selection of candidate tokens, allowing for greater variation in the generated text. This observation remains consistent across both the OPT-1.3B and Llama3-8B models.

\autoref{tab:impact_of_alpha_opt} and \autoref{tab:impact_of_alpha_llama} analyze the performance of smoothing attacks against different watermarking algorithms under varying values of \( \alpha \). In this attack, the weight assigned to the top-\( K \) tokens from the watermarked model is given by \( c^\alpha \), while the weight for the top-\( K \) tokens from the reference model is \( 1 - c^\alpha \), where \( c \) is a confidence score between 0 and 1. A larger \( \alpha \) shifts the token selection preference toward the reference model, making the generated text more aligned with it. Conversely, a smaller \( \alpha \) biases the attack toward the watermarked model, producing text that more closely resembles the watermarked output. As \( \alpha \) increases, the true positive rate (TPR) decreases, leading to a higher watermark removal rate—an effect consistently observed across all watermarking methods for both the OPT-1.3B and Llama3-8B models.

In terms of text quality, when \( \alpha \) is lower, the generated text is more influenced by the watermarked model, which generally exhibits higher quality than the reference model. Consequently, decreasing \( \alpha \) can improve text quality. This provides an adversary with a way to adjust \( \alpha \) to balance watermark removal and text quality preservation.

\subsection{Text example}
\label{appendix:text_example}
\autoref{tab:gumbel_example} presents text generated by the Gumbel sampling algorithm and the smoothing attack. We observe that, although the perplexity of the watermarked text is significantly lower than that of the text from the smoothing attack, this is primarily due to repetition in the generated text. This behavior may stem from the deterministic nature of Gumbel sampling, which can lead to less diverse outputs.
\input{data/tab_text_example_gumbel.tex}

\subsection{Impact of model size}
\label{appendix:impact_of_model_size}
\autoref{tab:impact_model_size} presents the performance of the smoothing attack across different watermarking algorithms and varying sizes of OPT models. Perplexity (PPL) is computed with respect to the OPT-30B model, while the reference model remains consistent across all settings—the OPT-125M.

For unwatermarked models, the True Positive Rate (TPR) is consistently 0\%. In contrast, watermarked models achieve near-perfect TPR. However, the smoothing attack significantly reduces TPR across all model sizes, with its impact increasing as the model size grows—for instance, TPR drops to 0\% for the KGW watermark in the 30B model.

Watermarked models exhibit a notable increase in perplexity, indicating that watermarking impacts text fluency. The smoothing attack reduces perplexity, bringing it closer to unwatermarked levels, suggesting a partial recovery of fluency. Regarding diversity, the unwatermarked text demonstrates the highest variation, while watermarking constrains generation patterns, resulting in a noticeable drop in diversity. The smoothing attack further reduces diversity, primarily because tokens are sampled only from the top-K tokens of both the watermarked and reference models, limiting the range of possible candidates.

\input{data/tab_impact_model_size_opt_rotated.tex}

%% file: figure_scripts/fig_scatter_trade-off.tex
\begin{subfigure}[b]{0.9\linewidth}
\centering
\begin{tikzpicture}
\begin{axis}[
    width=0.9\linewidth,
    height=5.5cm,
    ylabel={TPR (in \%)},
    xmin=2, xmax=20,
    ymin=-5, ymax=105,
    grid=both,
    title={(a) No attack},
    legend style={
        font=\scriptsize,
        at={(1.02,1.0)},
        anchor=north west
    }
]

\def\shapeNone{*} %

\addplot[
    only marks,
    mark=\shapeNone,
    mark options={scale=1.5},
    color=black
]
coordinates {
    (11.39, 1)  %
    (3.47, 1)   %
    (12.26, 1)  %
};
\addlegendentry{Un-watermarked}

\addplot[
    only marks,
    mark=\shapeNone,
    mark options={scale=1.5},
    color=gray
]
coordinates {
    (19.57, 1)
    (4.40, 1)
    (16.02, 1)
};
\addlegendentry{Reference}

\addplot[
    only marks,
    mark=\shapeNone,
    mark options={scale=1.5},
    color=MedBlue
]
coordinates {
    (14.61, 100)
    (4.60, 99)
    (16.46, 100)
};
\addlegendentry{KGW}

\addplot[
    only marks,
    mark=\shapeNone,
    mark options={scale=1.5},
    color=DarkBlue
]
coordinates {
    (14.99, 100)
    (4.61, 99)
    (15.41, 100)
};
\addlegendentry{Unigram}

\addplot[
    only marks,
    mark=\shapeNone,
    mark options={scale=1.5},
    color=Pink
]
coordinates {
    (7.12, 100)
    (4.83, 99)
    (6.94, 100)
};
\addlegendentry{SynthID}

\addplot[
    only marks,
    mark=\shapeNone,
    mark options={scale=1.5},
    color=Purple
]
coordinates {
    (13.73, 100)
    (4.03, 84)
    (14.34, 100)
};
\addlegendentry{DIP}

\addplot[
    only marks,
    mark=\shapeNone,
    mark options={scale=1.5},
    color=Green
]
coordinates {
    (13.61, 100)
    (4.02, 84)
    (14.64, 100)
};
\addlegendentry{Unbiased}

\addplot[
    only marks,
    mark=\shapeNone,
    mark options={scale=1.5},
    color=Orange
]
coordinates {
    (11.65, 99)
    (4.38, 83)
    (11.93, 86)
};
\addlegendentry{UPV}

\addplot[
    only marks,
    mark=\shapeNone,
    mark options={scale=1.5},
    color=Yellow
]
coordinates {
    (15.23, 100)
    (4.56, 100)
    (16.31, 100)
};
\addlegendentry{EWD}

\addplot[
    only marks,
    mark=\shapeNone,
    mark options={scale=1.5},
    color=Red
]
coordinates {
    (14.36, 100)
    (4.53, 99)
    (15.89, 100)
};
\addlegendentry{SWEET}

\end{axis}
\end{tikzpicture}
\end{subfigure}
\hfill
\begin{subfigure}[b]{0.9\linewidth}
\centering
\begin{tikzpicture}
\begin{axis}[
    width=0.9\linewidth,
    height=5.5cm,
    ylabel={TPR (in \%)},
    xmin=2, xmax=20,
    ymin=-5, ymax=105,
    grid=both,
    title={(b) Paraphrasing},
    legend style={
        font=\scriptsize,
        at={(1.02,1.0)},
        anchor=north west
    }
]

\def\shapePara{triangle*} %

\addplot[
    only marks,
    mark=\shapePara,
    mark options={scale=1.5},
    color=MedBlue
]
coordinates {
    (14.82, 3)
    (5.35, 2)
    (10.45, 2)
};
\addlegendentry{KGW}

\addplot[
    only marks,
    mark=\shapePara,
    mark options={scale=1.5},
    color=DarkBlue
]
coordinates {
    (14.51, 53)
    (5.60, 54)
    (10.40, 5)
};
\addlegendentry{Unigram}

\addplot[
    only marks,
    mark=\shapePara,
    mark options={scale=1.5},
    color=Pink
]
coordinates {
    (10.57, 1)
    (5.62, 1)
    (6.90, 1)
};
\addlegendentry{SynthID}

\addplot[
    only marks,
    mark=\shapePara,
    mark options={scale=1.5},
    color=Purple
]
coordinates {
    (13.95, 0)
    (5.25, 0)
    (10.10, 2)
};
\addlegendentry{DIP}

\addplot[
    only marks,
    mark=\shapePara,
    mark options={scale=1.5},
    color=Green
]
coordinates {
    (14.45, 3)
    (5.36, 2)
    (9.97, 1)
};
\addlegendentry{Unbiased}

\addplot[
    only marks,
    mark=\shapePara,
    mark options={scale=1.5},
    color=Orange
]
coordinates {
    (13.73, 34)
    (5.43, 2)
    (9.03, 2)
};
\addlegendentry{UPV}

\addplot[
    only marks,
    mark=\shapePara,
    mark options={scale=1.5},
    color=Yellow
]
coordinates {
    (14.95, 0)
    (5.73, 7)
    (10.18, 1)
};
\addlegendentry{EWD}

\addplot[
    only marks,
    mark=\shapePara,
    mark options={scale=1.5},
    color=Red
]
coordinates {
    (14.57, 0)
    (5.64, 14)
    (10.18, 4)
};
\addlegendentry{SWEET}

\end{axis}
\end{tikzpicture}
\end{subfigure}
\hfill
\begin{subfigure}[b]{0.9\linewidth}
\centering
\begin{tikzpicture}
\begin{axis}[
    width=0.9\linewidth,
    height=5.5cm,
    xlabel={Perplexity (PPL)},
    ylabel={TPR (in \%)},
    xmin=2, xmax=20,
    ymin=-5, ymax=105,
    grid=both,
    title={(c) Smoothing},
    legend style={
        font=\scriptsize,
        at={(1.02,1.0)},
        anchor=north west
    }
]

\def\shapeSmooth{square*} %

\addplot[
    only marks,
    mark=\shapeSmooth,
    mark options={scale=1.5},
    color=MedBlue
]
coordinates {
    (9.57, 0)
    (3.20, 2)
    (8.02, 0)
};
\addlegendentry{KGW}

\addplot[
    only marks,
    mark=\shapeSmooth,
    mark options={scale=1.5},
    color=DarkBlue
]
coordinates {
    (9.44, 5)
    (3.10, 24)
    (7.77, 1)
};
\addlegendentry{Unigram}

\addplot[
    only marks,
    mark=\shapeSmooth,
    mark options={scale=1.5},
    color=Pink
]
coordinates {
    (10.40, 0)
    (3.40, 0)
    (10.21, 0)
};
\addlegendentry{SynthID}

\addplot[
    only marks,
    mark=\shapeSmooth,
    mark options={scale=1.5},
    color=Purple
]
coordinates {
    (9.34, 6)
    (3.17, 6)
    (7.62, 11)
};
\addlegendentry{DIP}

\addplot[
    only marks,
    mark=\shapeSmooth,
    mark options={scale=1.5},
    color=Green
]
coordinates {
    (9.19, 27)
    (3.17, 5)
    (7.68, 5)
};
\addlegendentry{Unbiased}

\addplot[
    only marks,
    mark=\shapeSmooth,
    mark options={scale=1.5},
    color=Orange
]
coordinates {
    (10.01, 20)
    (3.12, 1)
    (8.16, 0)
};
\addlegendentry{UPV}

\addplot[
    only marks,
    mark=\shapeSmooth,
    mark options={scale=1.5},
    color=Yellow
]
coordinates {
    (9.93, 0)
    (3.13, 3)
    (7.82, 0)
};
\addlegendentry{EWD}

\addplot[
    only marks,
    mark=\shapeSmooth,
    mark options={scale=1.5},
    color=Red
]
coordinates {
    (9.59, 0)
    (3.09, 4)
    (7.85, 0)
};
\addlegendentry{SWEET}

\end{axis}
\end{tikzpicture}
\end{subfigure}

%% file: figure_scripts/fig_boxplot_ppl.tex
\begin{tikzpicture}[rotate=-90,transform shape]
    \begin{axis}[title={\small KGW}, %
        title style={
            at={(0,0.5)}, %
            anchor=south, %
            rotate=90 %
        },  scale=1, width=0.24\linewidth,height=0.24\linewidth, ymin=0,ymax=35,yticklabel= { \pgfmathprintnumber\tick}, ylabel={},ylabel near ticks, yticklabel pos=right, enlarge x limits=0.25, x tick label style={font=\small}, y tick label style={font=\small}, xticklabels={Unwatermarked, Watermarked, Reference, Paraphrasing, Smoothing }, xtick={0,1,2,3,4},xmin=0,xmax=4,xticklabel style={rotate=90, anchor=east},yticklabel style={rotate=90, anchor=north}]
        \addplot [box plot median] table {data/opt/KGW_ppl.dat};
        \addplot [style={fill=bblue}, restrict x to domain=0:0,box plot box] table {data/opt/KGW_ppl.dat};
        \addplot [style={fill=bgreen}, restrict x to domain=1:3,box plot box] table {data/opt/KGW_ppl.dat};
        \addplot [style={fill=bred}, restrict x to domain=4:4,box plot box] table {data/opt/KGW_ppl.dat};
        \addplot [box plot top whisker] table {data/opt/KGW_ppl.dat};
        \addplot [box plot bottom whisker] table {data/opt/KGW_ppl.dat};
    \end{axis}
\end{tikzpicture}
\begin{tikzpicture}[rotate=-90,transform shape]
    \begin{axis}[title={\small Unigram}, %
        title style={
            at={(0,0.5)}, %
            anchor=south, %
            rotate=90 %
        }, scale=1.0, width=0.24\linewidth,height=0.24\linewidth, ymin=0,ymax=35,yticklabel= { \pgfmathprintnumber\tick}, ylabel={},ylabel near ticks, yticklabel pos=right, enlarge x limits=0.25, x tick label style={font=\small}, y tick label style={font=\small}, xticklabels={}, xtick={0,1,2,3,4},xmin=0,xmax=4,xticklabel style={rotate=90, anchor=east},yticklabel style={rotate=90, anchor=north}]
        \addplot [box plot median] table {data/opt/Unigram_ppl.dat};
        \addplot [style={fill=bblue}, restrict x to domain=0:0,box plot box] table {data/opt/Unigram_ppl.dat};
        \addplot [style={fill=bgreen}, restrict x to domain=1:3,box plot box] table {data/opt/Unigram_ppl.dat};
        \addplot [style={fill=bred}, restrict x to domain=4:4,box plot box] table {data/opt/Unigram_ppl.dat};
        \addplot [box plot top whisker] table {data/opt/Unigram_ppl.dat};
        \addplot [box plot bottom whisker] table {data/opt/Unigram_ppl.dat};
    \end{axis}
\end{tikzpicture}
\begin{tikzpicture}[rotate=-90,transform shape]
    \begin{axis}[title={\small SynthID}, %
        title style={
            at={(0,0.5)}, %
            anchor=south, %
            rotate=90 %
        }, scale=1, width=0.24\linewidth,height=0.24\linewidth, ymin=0,ymax=35,yticklabel= {\pgfmathprintnumber\tick}, ylabel={},ylabel near ticks, yticklabel pos=right, enlarge x limits=0.25, x tick label style={font=\small}, y tick label style={font=\small}, xticklabels={}, xtick={0,1,2,3,4},xmin=0,xmax=4,xticklabel style={rotate=90, anchor=east},yticklabel style={rotate=90, anchor=north}]
        \addplot [box plot median] table {data/opt/SynthID_ppl.dat};
        \addplot [style={fill=bblue}, restrict x to domain=0:0,box plot box] table {data/opt/SynthID_ppl.dat};
        \addplot [style={fill=bgreen}, restrict x to domain=1:3,box plot box] table {data/opt/SynthID_ppl.dat};
        \addplot [style={fill=bred}, restrict x to domain=4:4,box plot box] table {data/opt/SynthID_ppl.dat};
        \addplot [box plot top whisker] table {data/opt/SynthID_ppl.dat};
        \addplot [box plot bottom whisker] table {data/opt/SynthID_ppl.dat};
    \end{axis}
\end{tikzpicture}
\begin{tikzpicture}[rotate=-90,transform shape]
    \begin{axis}[title={\small DIP}, %
        title style={
            at={(0,0.5)}, %
            anchor=south, %
            rotate=90 %
        }, scale=1, width=0.24\linewidth,height=0.24\linewidth, ymin=0,ymax=35,yticklabel= {\pgfmathprintnumber\tick}, ylabel={},ylabel near ticks, yticklabel pos=right, enlarge x limits=0.25, x tick label style={font=\small}, y tick label style={font=\small}, xticklabels={}, xtick={0,1,2,3,4},xmin=0,xmax=4,xticklabel style={rotate=90, anchor=east},yticklabel style={rotate=90, anchor=north}]
        \addplot [box plot median] table {data/opt/DIP_ppl.dat};
        \addplot [style={fill=bblue}, restrict x to domain=0:0,box plot box] table {data/opt/DIP_ppl.dat};
        \addplot [style={fill=bgreen}, restrict x to domain=1:3,box plot box] table {data/opt/DIP_ppl.dat};
        \addplot [style={fill=bred}, restrict x to domain=4:4,box plot box] table {data/opt/DIP_ppl.dat};
        \addplot [box plot top whisker] table {data/opt/DIP_ppl.dat};
        \addplot [box plot bottom whisker] table {data/opt/DIP_ppl.dat};
    \end{axis}
\end{tikzpicture}
\begin{tikzpicture}[rotate=-90,transform shape]
    \begin{axis}[title={\small Unbiased}, %
        title style={
            at={(0,0.5)}, %
            anchor=south, %
            rotate=90 %
        }, scale=1, width=0.24\linewidth,height=0.24\linewidth, ymin=0,ymax=35,yticklabel= {\pgfmathprintnumber\tick}, ylabel={},ylabel near ticks, yticklabel pos=right, enlarge x limits=0.25, x tick label style={font=\small}, xticklabels={}, y tick label style={font=\small},  xmin=0,xmax=4,xticklabel style={rotate=90, anchor=east},yticklabel style={rotate=90, anchor=north}]
        \addplot [box plot median] table {data/opt/Unbiased_ppl.dat};
        \addplot [style={fill=bblue}, restrict x to domain=0:0,box plot box] table {data/opt/Unbiased_ppl.dat};
        \addplot [style={fill=bgreen}, restrict x to domain=1:3,box plot box] table {data/opt/Unbiased_ppl.dat};
        \addplot [style={fill=bred}, restrict x to domain=4:4,box plot box] table {data/opt/Unbiased_ppl.dat};
        \addplot [box plot top whisker] table {data/opt/Unbiased_ppl.dat};
        \addplot [box plot bottom whisker] table {data/opt/Unbiased_ppl.dat};
    \end{axis}
\end{tikzpicture}
\begin{tikzpicture}[rotate=-90,transform shape]
    \begin{axis}[title={\small UPV}, %
        title style={
            at={(0,0.5)}, %
            anchor=south, %
            rotate=90 %
        }, scale=1, width=0.24\linewidth,height=0.24\linewidth, ymin=0,ymax=35,yticklabel= {\pgfmathprintnumber\tick}, ylabel={\small Perplexity},ylabel near ticks, yticklabel pos=right, enlarge x limits=0.25, x tick label style={font=\small}, y tick label style={font=\small}, xtick={0,1,2,3,4}, xticklabels={Unwatermarked, Watermarked, Reference, Paraphrasing, Smoothing }, xtick={0,1,2,3,4},xmin=0,xmax=4,xticklabel style={rotate=90, anchor=east},yticklabel style={rotate=90, anchor=north}]
        \addplot [box plot median] table {data/opt/UPV_ppl.dat};
        \addplot [style={fill=bblue}, restrict x to domain=0:0,box plot box] table {data/opt/UPV_ppl.dat};
        \addplot [style={fill=bgreen}, restrict x to domain=1:3,box plot box] table {data/opt/UPV_ppl.dat};
        \addplot [style={fill=bred}, restrict x to domain=4:4,box plot box] table {data/opt/UPV_ppl.dat};
        \addplot [box plot top whisker] table {data/opt/UPV_ppl.dat};
        \addplot [box plot bottom whisker] table {data/opt/UPV_ppl.dat};
    \end{axis}
\end{tikzpicture}
\begin{tikzpicture}[rotate=-90,transform shape]
    \begin{axis}[title={\small EWD}, %
        title style={
            at={(0,0.5)}, %
            anchor=south, %
            rotate=90 %
        }, scale=1, width=0.24\linewidth,height=0.24\linewidth, ymin=0,ymax=35,yticklabel= {\pgfmathprintnumber\tick}, ylabel={\small Perplexity},ylabel near ticks, yticklabel pos=right, enlarge x limits=0.25, x tick label style={font=\small}, y tick label style={font=\small}, xticklabels={}, xtick={0,1,2,3,4},xmin=0,xmax=4,xticklabel style={rotate=90, anchor=east},yticklabel style={rotate=90, anchor=north}]
        \addplot [box plot median] table {data/opt/EWD_ppl.dat};
        \addplot [style={fill=bblue}, restrict x to domain=0:0,box plot box] table {data/opt/EWD_ppl.dat};
        \addplot [style={fill=bgreen}, restrict x to domain=1:3,box plot box] table {data/opt/EWD_ppl.dat};
        \addplot [style={fill=bred}, restrict x to domain=4:4,box plot box] table {data/opt/EWD_ppl.dat};
        \addplot [box plot top whisker] table {data/opt/EWD_ppl.dat};
        \addplot [box plot bottom whisker] table {data/opt/EWD_ppl.dat};
    \end{axis}
\end{tikzpicture}
\begin{tikzpicture}[rotate=-90,transform shape]
    \begin{axis}[title={\small SWEET}, %
        title style={
            at={(0,0.5)}, %
            anchor=south, %
            rotate=90 %
        }, scale=1, width=0.24\linewidth,height=0.24\linewidth, ymin=0,ymax=35,yticklabel= {\pgfmathprintnumber\tick}, ylabel={\small Perplexity},ylabel near ticks, yticklabel pos=right, enlarge x limits=0.25, x tick label style={font=\small}, y tick label style={font=\small}, xticklabels={}, xtick={0,1,2,3,4},xmin=0,xmax=4,xticklabel style={rotate=90, anchor=east},yticklabel style={rotate=90, anchor=north}]
        \addplot [box plot median] table {data/opt/SWEET_ppl.dat};
        \addplot [style={fill=bblue}, restrict x to domain=0:0,box plot box] table {data/opt/SWEET_ppl.dat};
        \addplot [style={fill=bgreen}, restrict x to domain=1:3,box plot box] table {data/opt/SWEET_ppl.dat};
        \addplot [style={fill=bred}, restrict x to domain=4:4,box plot box] table {data/opt/SWEET_ppl.dat};
        \addplot [box plot top whisker] table {data/opt/SWEET_ppl.dat};
        \addplot [box plot bottom whisker] table {data/opt/SWEET_ppl.dat};
    \end{axis}
\end{tikzpicture}
\begin{tikzpicture}[rotate=-90,transform shape]
    \begin{axis}[title={\small XSIR}, %
        title style={
            at={(0,0.5)}, %
            anchor=south, %
            rotate=90 %
        }, scale=1, width=0.24\linewidth,height=0.24\linewidth, ymin=0,ymax=35,yticklabel= {\pgfmathprintnumber\tick}, ylabel={\small Perplexity},ylabel near ticks, yticklabel pos=right, enlarge x limits=0.25, x tick label style={font=\small}, y tick label style={font=\small}, xticklabels={}, xtick={0,1,2,3,4},xmin=0,xmax=4,xticklabel style={rotate=90, anchor=east},yticklabel style={rotate=90, anchor=north}]
        \addplot [box plot median] table {data/opt/XSIR_ppl.dat};
        \addplot [style={fill=bblue}, restrict x to domain=0:0,box plot box] table {data/opt/XSIR_ppl.dat};
        \addplot [style={fill=bgreen}, restrict x to domain=1:3,box plot box] table {data/opt/XSIR_ppl.dat};
        \addplot [style={fill=bred}, restrict x to domain=4:4,box plot box] table {data/opt/XSIR_ppl.dat};
        \addplot [box plot top whisker] table {data/opt/XSIR_ppl.dat};
        \addplot [box plot bottom whisker] table {data/opt/XSIR_ppl.dat};
    \end{axis}
\end{tikzpicture}
\begin{tikzpicture}[rotate=-90,transform shape]
    \begin{axis}[title={\small Gumbel}, %
        title style={
            at={(0,0.5)}, %
            anchor=south, %
            rotate=90 %
        }, scale=1, width=0.24\linewidth,height=0.24\linewidth, ymin=0,ymax=35,yticklabel= {\pgfmathprintnumber\tick}, ylabel={\small Perplexity},ylabel near ticks, yticklabel pos=right, enlarge x limits=0.25, x tick label style={font=\small}, y tick label style={font=\small}, xticklabels={}, xtick={0,1,2,3,4},xmin=0,xmax=4,xticklabel style={rotate=90, anchor=east},yticklabel style={rotate=90, anchor=north}]
        \addplot [box plot median] table {data/opt/Gumble_ppl.dat};
        \addplot [style={fill=bblue}, restrict x to domain=0:0,box plot box] table {data/opt/Gumble_ppl.dat};
        \addplot [style={fill=bgreen}, restrict x to domain=1:3,box plot box] table {data/opt/Gumble_ppl.dat};
        \addplot [style={fill=bred}, restrict x to domain=4:4,box plot box] table {data/opt/Gumble_ppl.dat};
        \addplot [box plot top whisker] table {data/opt/Gumble_ppl.dat};
        \addplot [box plot bottom whisker] table {data/opt/Gumble_ppl.dat};
    \end{axis}
\end{tikzpicture}

%% file: figure_scripts/fig_boxplot_diversity.tex
\begin{tikzpicture}[rotate=-90,transform shape]
    \begin{axis}[title={\small KGW}, %
        title style={
            at={(0,0.5)}, %
            anchor=south, %
            rotate=90 %
        },  scale=1, width=0.24\linewidth,height=0.24\linewidth, ymin=0,ymax=25,yticklabel= { \pgfmathprintnumber\tick}, ylabel={},ylabel near ticks, yticklabel pos=right, enlarge x limits=0.25, x tick label style={font=\small}, y tick label style={font=\small}, xticklabels={Unwatermarked, Watermarked, Reference, Paraphrasing, Smoothing }, xtick={0,1,2,3,4},xmin=0,xmax=4,xticklabel style={rotate=90, anchor=east},yticklabel style={rotate=90, anchor=north}]
        \addplot [box plot median] table {data/opt/KGW_diversity.dat};
        \addplot [style={fill=bblue}, restrict x to domain=0:0,box plot box] table {data/opt/KGW_diversity.dat};
        \addplot [style={fill=bgreen}, restrict x to domain=1:3,box plot box] table {data/opt/KGW_diversity.dat};
        \addplot [style={fill=bred}, restrict x to domain=4:4,box plot box] table {data/opt/KGW_diversity.dat};
        \addplot [box plot top whisker] table {data/opt/KGW_diversity.dat};
        \addplot [box plot bottom whisker] table {data/opt/KGW_diversity.dat};
    \end{axis}
\end{tikzpicture}
\begin{tikzpicture}[rotate=-90,transform shape]
    \begin{axis}[title={\small Unigram}, %
        title style={
            at={(0,0.5)}, %
            anchor=south, %
            rotate=90 %
        }, scale=1.0, width=0.24\linewidth,height=0.24\linewidth, ymin=0,ymax=25,yticklabel= { \pgfmathprintnumber\tick}, ylabel={},ylabel near ticks, yticklabel pos=right, enlarge x limits=0.25, x tick label style={font=\small}, y tick label style={font=\small}, xticklabels={}, xtick={0,1,2,3,4},xmin=0,xmax=4,xticklabel style={rotate=90, anchor=east},yticklabel style={rotate=90, anchor=north}]
        \addplot [box plot median] table {data/opt/Unigram_diversity.dat};
        \addplot [style={fill=bblue}, restrict x to domain=0:0,box plot box] table {data/opt/Unigram_diversity.dat};
        \addplot [style={fill=bgreen}, restrict x to domain=1:3,box plot box] table {data/opt/Unigram_diversity.dat};
        \addplot [style={fill=bred}, restrict x to domain=4:4,box plot box] table {data/opt/Unigram_diversity.dat};
        \addplot [box plot top whisker] table {data/opt/Unigram_diversity.dat};
        \addplot [box plot bottom whisker] table {data/opt/Unigram_diversity.dat};
    \end{axis}
\end{tikzpicture}
\begin{tikzpicture}[rotate=-90,transform shape]
    \begin{axis}[title={\small SynthID}, %
        title style={
            at={(0,0.5)}, %
            anchor=south, %
            rotate=90 %
        }, scale=1, width=0.24\linewidth,height=0.24\linewidth, ymin=0,ymax=25,yticklabel= {\pgfmathprintnumber\tick}, ylabel={},ylabel near ticks, yticklabel pos=right, enlarge x limits=0.25, x tick label style={font=\small}, y tick label style={font=\small}, xticklabels={}, xtick={0,1,2,3,4},xmin=0,xmax=4,xticklabel style={rotate=90, anchor=east},yticklabel style={rotate=90, anchor=north}]
        \addplot [box plot median] table {data/opt/SynthID_diversity.dat};
        \addplot [style={fill=bblue}, restrict x to domain=0:0,box plot box] table {data/opt/SynthID_diversity.dat};
        \addplot [style={fill=bgreen}, restrict x to domain=1:3,box plot box] table {data/opt/SynthID_diversity.dat};
        \addplot [style={fill=bred}, restrict x to domain=4:4,box plot box] table {data/opt/SynthID_diversity.dat};
        \addplot [box plot top whisker] table {data/opt/SynthID_diversity.dat};
        \addplot [box plot bottom whisker] table {data/opt/SynthID_diversity.dat};
    \end{axis}
\end{tikzpicture}
\begin{tikzpicture}[rotate=-90,transform shape]
    \begin{axis}[title={\small DIP}, %
        title style={
            at={(0,0.5)}, %
            anchor=south, %
            rotate=90 %
        }, scale=1, width=0.24\linewidth,height=0.24\linewidth, ymin=0,ymax=25,yticklabel= {\pgfmathprintnumber\tick}, ylabel={},ylabel near ticks, yticklabel pos=right, enlarge x limits=0.25, x tick label style={font=\small}, y tick label style={font=\small}, xticklabels={}, xtick={0,1,2,3,4},xmin=0,xmax=4,xticklabel style={rotate=90, anchor=east},yticklabel style={rotate=90, anchor=north}]
        \addplot [box plot median] table {data/opt/DIP_diversity.dat};
        \addplot [style={fill=bblue}, restrict x to domain=0:0,box plot box] table {data/opt/DIP_diversity.dat};
        \addplot [style={fill=bgreen}, restrict x to domain=1:3,box plot box] table {data/opt/DIP_diversity.dat};
        \addplot [style={fill=bred}, restrict x to domain=4:4,box plot box] table {data/opt/DIP_diversity.dat};
        \addplot [box plot top whisker] table {data/opt/DIP_diversity.dat};
        \addplot [box plot bottom whisker] table {data/opt/DIP_diversity.dat};
    \end{axis}
\end{tikzpicture}
\begin{tikzpicture}[rotate=-90,transform shape]
    \begin{axis}[title={\small Unbiased}, %
        title style={
            at={(0,0.5)}, %
            anchor=south, %
            rotate=90 %
        }, scale=1, width=0.24\linewidth,height=0.24\linewidth, ymin=0,ymax=25,yticklabel= {\pgfmathprintnumber\tick}, ylabel={},ylabel near ticks, yticklabel pos=right, enlarge x limits=0.25, x tick label style={font=\small}, xticklabels={}, y tick label style={font=\small},  xmin=0,xmax=4,xticklabel style={rotate=90, anchor=east},yticklabel style={rotate=90, anchor=north}]
        \addplot [box plot median] table {data/opt/Unbiased_diversity.dat};
        \addplot [style={fill=bblue}, restrict x to domain=0:0,box plot box] table {data/opt/Unbiased_diversity.dat};
        \addplot [style={fill=bgreen}, restrict x to domain=1:3,box plot box] table {data/opt/Unbiased_diversity.dat};
        \addplot [style={fill=bred}, restrict x to domain=4:4,box plot box] table {data/opt/Unbiased_diversity.dat};
        \addplot [box plot top whisker] table {data/opt/Unbiased_diversity.dat};
        \addplot [box plot bottom whisker] table {data/opt/Unbiased_diversity.dat};
    \end{axis}
\end{tikzpicture}
\begin{tikzpicture}[rotate=-90,transform shape]
    \begin{axis}[title={\small UPV}, %
        title style={
            at={(0,0.5)}, %
            anchor=south, %
            rotate=90 %
        }, scale=1, width=0.24\linewidth,height=0.24\linewidth, ymin=0,ymax=25,yticklabel= {\pgfmathprintnumber\tick}, ylabel={\small Diversity},ylabel near ticks, yticklabel pos=right, enlarge x limits=0.25, x tick label style={font=\small}, y tick label style={font=\small}, xtick={0,1,2,3,4}, xticklabels={Unwatermarked, Watermarked, Reference, Paraphrasing, Smoothing }, xtick={0,1,2,3,4},xmin=0,xmax=4,xticklabel style={rotate=90, anchor=east},yticklabel style={rotate=90, anchor=north}]
        \addplot [box plot median] table {data/opt/UPV_diversity.dat};
        \addplot [style={fill=bblue}, restrict x to domain=0:0,box plot box] table {data/opt/UPV_diversity.dat};
        \addplot [style={fill=bgreen}, restrict x to domain=1:3,box plot box] table {data/opt/UPV_diversity.dat};
        \addplot [style={fill=bred}, restrict x to domain=4:4,box plot box] table {data/opt/UPV_diversity.dat};
        \addplot [box plot top whisker] table {data/opt/UPV_diversity.dat};
        \addplot [box plot bottom whisker] table {data/opt/UPV_diversity.dat};
    \end{axis}
\end{tikzpicture}
\begin{tikzpicture}[rotate=-90,transform shape]
    \begin{axis}[title={\small EWD}, %
        title style={
            at={(0,0.5)}, %
            anchor=south, %
            rotate=90 %
        }, scale=1, width=0.24\linewidth,height=0.24\linewidth, ymin=0,ymax=25,yticklabel= {\pgfmathprintnumber\tick}, ylabel={\small Diversity},ylabel near ticks, yticklabel pos=right, enlarge x limits=0.25, x tick label style={font=\small}, y tick label style={font=\small}, xticklabels={}, xtick={0,1,2,3,4},xmin=0,xmax=4,xticklabel style={rotate=90, anchor=east},yticklabel style={rotate=90, anchor=north}]
        \addplot [box plot median] table {data/opt/EWD_diversity.dat};
        \addplot [style={fill=bblue}, restrict x to domain=0:0,box plot box] table {data/opt/EWD_diversity.dat};
        \addplot [style={fill=bgreen}, restrict x to domain=1:3,box plot box] table {data/opt/EWD_diversity.dat};
        \addplot [style={fill=bred}, restrict x to domain=4:4,box plot box] table {data/opt/EWD_diversity.dat};
        \addplot [box plot top whisker] table {data/opt/EWD_diversity.dat};
        \addplot [box plot bottom whisker] table {data/opt/EWD_diversity.dat};
    \end{axis}
\end{tikzpicture}
\begin{tikzpicture}[rotate=-90,transform shape]
    \begin{axis}[title={\small SWEET}, %
        title style={
            at={(0,0.5)}, %
            anchor=south, %
            rotate=90 %
        }, scale=1, width=0.24\linewidth,height=0.24\linewidth, ymin=0,ymax=25,yticklabel= {\pgfmathprintnumber\tick}, ylabel={\small Diversity},ylabel near ticks, yticklabel pos=right, enlarge x limits=0.25, x tick label style={font=\small}, y tick label style={font=\small}, xticklabels={}, xtick={0,1,2,3,4},xmin=0,xmax=4,xticklabel style={rotate=90, anchor=east},yticklabel style={rotate=90, anchor=north}]
        \addplot [box plot median] table {data/opt/SWEET_diversity.dat};
        \addplot [style={fill=bblue}, restrict x to domain=0:0,box plot box] table {data/opt/SWEET_diversity.dat};
        \addplot [style={fill=bgreen}, restrict x to domain=1:3,box plot box] table {data/opt/SWEET_diversity.dat};
        \addplot [style={fill=bred}, restrict x to domain=4:4,box plot box] table {data/opt/SWEET_diversity.dat};
        \addplot [box plot top whisker] table {data/opt/SWEET_diversity.dat};
        \addplot [box plot bottom whisker] table {data/opt/SWEET_diversity.dat};
    \end{axis}
\end{tikzpicture}
\begin{tikzpicture}[rotate=-90,transform shape]
    \begin{axis}[title={\small XSIR}, %
        title style={
            at={(0,0.5)}, %
            anchor=south, %
            rotate=90 %
        }, scale=1, width=0.24\linewidth,height=0.24\linewidth, ymin=0,ymax=25,yticklabel= {\pgfmathprintnumber\tick}, ylabel={\small Diversity},ylabel near ticks, yticklabel pos=right, enlarge x limits=0.25, x tick label style={font=\small}, y tick label style={font=\small}, xticklabels={}, xtick={0,1,2,3,4},xmin=0,xmax=4,xticklabel style={rotate=90, anchor=east},yticklabel style={rotate=90, anchor=north}]
        \addplot [box plot median] table {data/opt/XSIR_diversity.dat};
        \addplot [style={fill=bblue}, restrict x to domain=0:0,box plot box] table {data/opt/XSIR_diversity.dat};
        \addplot [style={fill=bgreen}, restrict x to domain=1:3,box plot box] table {data/opt/XSIR_diversity.dat};
        \addplot [style={fill=bred}, restrict x to domain=4:4,box plot box] table {data/opt/XSIR_diversity.dat};
        \addplot [box plot top whisker] table {data/opt/XSIR_diversity.dat};
        \addplot [box plot bottom whisker] table {data/opt/XSIR_diversity.dat};
    \end{axis}
\end{tikzpicture}
\begin{tikzpicture}[rotate=-90,transform shape]
    \begin{axis}[title={\small Gumbel}, %
        title style={
            at={(0,0.5)}, %
            anchor=south, %
            rotate=90 %
        }, scale=1, width=0.24\linewidth,height=0.24\linewidth, ymin=0,ymax=25,yticklabel= {\pgfmathprintnumber\tick}, ylabel={\small Diversity},ylabel near ticks, yticklabel pos=right, enlarge x limits=0.25, x tick label style={font=\small}, y tick label style={font=\small}, xticklabels={}, xtick={0,1,2,3,4},xmin=0,xmax=4,xticklabel style={rotate=90, anchor=east},yticklabel style={rotate=90, anchor=north}]
        \addplot [box plot median] table {data/opt/Gumble_diversity.dat};
        \addplot [style={fill=bblue}, restrict x to domain=0:0,box plot box] table {data/opt/Gumble_diversity.dat};
        \addplot [style={fill=bgreen}, restrict x to domain=1:3,box plot box] table {data/opt/Gumble_diversity.dat};
        \addplot [style={fill=bred}, restrict x to domain=4:4,box plot box] table {data/opt/Gumble_diversity.dat};
        \addplot [box plot top whisker] table {data/opt/Gumble_diversity.dat};
        \addplot [box plot bottom whisker] table {data/opt/Gumble_diversity.dat};
    \end{axis}
\end{tikzpicture}

%% file: figure_scripts/fig_heatmap_psp.tex
\begin{tikzpicture}[scale=0.8]
\node[font=\bfseries, align=center] at (2.5, 0) {KGW};
\foreach \y [count=\n] in {
{100, 55.9, 54.9, 45.9, 55.0},
{55.9, 100, 53.7, 44.6, 53.3},
{54.9, 53.7, 100, 43.3, 55.9},
{45.9, 44.6, 43.3, 100, 43.4},
{55.0, 53.3, 55.9, 43.4, 100},
} {
\foreach \x [count=\m] in \y {
    \node[fill=red!\x!white, minimum size=6mm, text=white] at (\m,-\n) {\x};
}
}
\foreach \a [count=\i] in {Unwatermarked, Watermarked, Reference, Paraphrasing, Smoothing} {
    \node[minimum size=6mm] at (-1,-\i) {\small \a};
}
\end{tikzpicture}
\begin{tikzpicture}[scale=0.8]
\node[font=\bfseries, align=center] at (2.5, 0) {Unigram};
\foreach \y [count=\n] in {
{100, 53.6, 54.0, 43.9, 55.4},
{53.6, 100, 52.0, 82.3, 54.2},
{54.0, 52.0, 100, 42.3, 56.1},
{43.9, 82.3, 42.3, 100, 43.2},
{55.4, 54.2, 56.1, 43.2, 100},
} {
\foreach \x [count=\m] in \y {
    \node[fill=red!\x!white, minimum size=6mm, text=white] at (\m,-\n) {\x};
}
}
\end{tikzpicture}
\begin{tikzpicture}[scale=0.8]
\node[font=\bfseries, align=center] at (2.5, 0) {SynthID};
\foreach \y [count=\n] in {
{100, 56.3, 54.0, 46.8, 63.0},
{56.3, 100, 54.1, 85.0, 56.3},
{54.0, 54.1, 100, 44.3, 54.0},
{46.8, 85.0, 44.3, 100, 47.5},
{63.0, 56.3, 54.0, 47.5, 100},
} {
\foreach \x [count=\m] in \y {
    \node[fill=red!\x!white, minimum size=6mm, text=white] at (\m,-\n) {\x};
}
}
\end{tikzpicture}
\begin{tikzpicture}[scale=0.8]
\node[font=\bfseries, align=center] at (2.5, 0) {DIP};
\foreach \y [count=\n] in {
{100, 57.0, 54.0, 47.2, 54.5},
{57.0, 100, 54.8, 83.7, 55.2},
{54.0, 54.8, 100, 44.6, 56.4},
{47.2, 83.7, 44.6, 100, 44.1},
{54.5, 55.2, 56.4, 44.1, 100},
} {
\foreach \x [count=\m] in \y {
    \node[fill=red!\x!white, minimum size=6mm, text=white] at (\m,-\n) {\x};
}
}
\foreach \a [count=\i] in {Unwatermarked, Watermarked, Reference, Paraphrasing, Smoothing} {
    \node[minimum size=6mm, font=\small, rotate=90] at (\i,-7) {\small \a};
}
\foreach \a [count=\i] in {Unwatermarked, Watermarked, Reference, Paraphrasing, Smoothing} {
    \node[minimum size=6mm] at (-1,-\i) {\small \a};
}
\end{tikzpicture}
\begin{tikzpicture}[scale=0.8]
\node[font=\bfseries, align=center] at (2.5, 0) {Unbiased};
\foreach \y [count=\n] in {
{100, 56.6, 54.0, 46.8, 54.5},
{56.6, 100, 54.0, 84.1, 53.5},
{54.0, 54.0, 100, 44.2, 55.7},
{46.8, 84.1, 44.2, 100, 42.5},
{54.5, 53.5, 55.7, 42.5, 100},
} {
\foreach \x [count=\m] in \y {
    \node[fill=red!\x!white, minimum size=6mm, text=white] at (\m,-\n) {\x};
}
}
\foreach \a [count=\i] in {Unwatermarked, Watermarked, Reference, Paraphrasing, Smoothing} {
    \node[minimum size=6mm, font=\small, rotate=90] at (\i,-7) {\small \a};
}
\end{tikzpicture}
\begin{tikzpicture}[scale=0.8]
    \node[font=\bfseries, align=center] at (2.5, 0) {XSIR};
    \foreach \y [count=\n] in {
        {100, 54.5, 54.0, 44.8, 54.8},
        {54.5, 100, 53.4, 83.7, 53.3},
        {54.0, 53.4, 100, 43.4, 56.5},
        {44.8, 83.7, 43.4, 100, 42.1},
        {54.8, 53.3, 56.5, 42.1, 100},
    } {
    \foreach \x [count=\m] in \y {
        \node[fill=red!\x!white, minimum size=6mm, text=white] at (\m,-\n) {\x};
    }
    }
    \foreach \a [count=\i] in {Unwatermarked, Watermarked, Reference, Paraphrasing, Smoothing} {
        \node[minimum size=6mm, font=\small, rotate=90] at (\i,-7) {\small \a};
    }
    \end{tikzpicture}

%% file: data/tab_impact_k_opt_rotated.tex
\begin{table*}[ht!]
\centering
\caption{Effect of $K$ on Smoothing Attack Performance (OPT-1.3B). Evaluation of the smoothing attack's effectiveness against different watermarking algorithms on the OPT-1.3B model, varying the number of top-$K$ tokens accessible to the attacker.}
\label{tab:impact_of_k_opt}
\begin{tabular}{lccc|ccc|ccc|ccc|ccc}
\toprule
K & \multicolumn{3}{c}{KGW} & \multicolumn{3}{c}{Unigram} & \multicolumn{3}{c}{SynthID} & \multicolumn{3}{c}{DIP} & \multicolumn{3}{c}{Unbiased} \\
\cmidrule(lr){2-4} \cmidrule(lr){5-7} \cmidrule(lr){8-10} \cmidrule(lr){11-13} \cmidrule(lr){14-16}
    & TPR & PPL & Div & TPR & PPL & Div & TPR & PPL & Div & TPR & PPL & Div & TPR & PPL & Div \\
\hline
1  & 9\% & 3.22 & 4.54 & 18\% & 3.21 & 4.62 & 0\% & 10.45 & 8.47 & 1\% & 3.36 & 4.57 & 7\% & 3.36 & 4.56 \\
3  & 0.0\% & 5.76 & 5.71 & 8.0\% & 5.9 & 5.68 & 0.0\% & 10.5 & 8.31 & 4.0\% & 5.58 & 5.66 & 14.0\% & 5.59 & 5.68 \\
5  & 2.0\% & 7.27 & 6.17 & 10.0\% & 7.46 & 6.11 & 0.0\% & 10.35 & 8.71 & 3.0\% & 6.97 & 6.23 & 19.0\% & 7.11 & 6.29 \\
7  & 1.0\% & 8.14 & 6.46 & 5.0\% & 8.48 & 6.55 & 0.0\% & 10.42 & 8.63 & 7.0\% & 7.97 & 6.46 & 29.0\% & 8.06 & 6.47 \\
10 & 0.0\% & 9.57 & 6.72 & 5.0\% & 9.44 & 6.73 & 0.0\% & 10.4 & 8.64 & 6.0\% & 9.34 & 6.84 & 27.0\% & 9.19 & 6.84 \\
\midrule\midrule
K & \multicolumn{3}{c}{XSIR} & \multicolumn{3}{c}{UPV} & \multicolumn{3}{c}{Gumbel} & \multicolumn{3}{c}{EWD} & \multicolumn{3}{c}{SWEET} \\
\cmidrule(lr){2-4} \cmidrule(lr){5-7} \cmidrule(lr){8-10} \cmidrule(lr){11-13} \cmidrule(lr){14-16}
    & TPR & PPL & Div & TPR & PPL & Div & TPR & PPL & Div & TPR & PPL & Div & TPR & PPL & Div \\
\hline
1  & 14\% & 3.31 & 4.5 & 22\% & 3.62 & 4.63 & 0\% & 20.8 & 8.2 & 1\% & 3.31 & 4.49 & 2\% & 3.41 & 4.57 \\
3  & 17.0\% & 5.69 & 5.52 & 14.0\% & 6.22 & 5.87 & 2.0\% & 21.72 & 8.47 & 0.0\% & 5.78 & 5.71 & 0.0\% & 5.64 & 5.75 \\
5  & 8.0\% & 6.8 & 6.04 & 16.0\% & 7.68 & 6.3 & 8.0\% & 20.3 & 8.23 & 0.0\% & 7.32 & 6.18 & 0.0\% & 7.15 & 6.23 \\
7  & 10.0\% & 8.26 & 6.48 & 7.0\% & 8.75 & 6.65 & 9.0\% & 21.15 & 8.15 & 0.0\% & 8.65 & 6.52 & 0.0\% & 8.47 & 6.45 \\
10 & 9.0\% & 9.47 & 6.75 & 20.0\% & 10.01 & 6.89 & 9.0\% & 19.25 & 8.3 & 0.0\% & 9.93 & 6.78 & 0.0\% & 9.59 & 6.72 \\
\bottomrule
\end{tabular}
\end{table*}

%% file: data/tab_impact_alpha_opt_rotated.tex
\begin{table*}[h!]
\centering
\caption{Effect of \( \alpha \) on Smoothing Attack Performance (OPT-1.3B). Evaluation of the smoothing attack's effectiveness against different watermarking algorithms on the OPT-1.3B model, varying the parameter \( \alpha \). A larger \( \alpha \) indicates that the attack relies more on the reference model's output, while a smaller \( \alpha \) means the attack is more influenced by the watermarked text.}
\label{tab:impact_of_alpha_opt}
\begin{tabular}{lccc|ccc|ccc|ccc|ccc}
\toprule
$\alpha$ & \multicolumn{3}{c}{KGW} & \multicolumn{3}{c}{Unigram} & \multicolumn{3}{c}{SynthID} & \multicolumn{3}{c}{DIP} & \multicolumn{3}{c}{Unbiased} \\
\cmidrule(lr){2-4} \cmidrule(lr){5-7} \cmidrule(lr){8-10} \cmidrule(lr){11-13} \cmidrule(lr){14-16}
    & TPR & PPL & Div & TPR & PPL & Div & TPR & PPL & Div & TPR & PPL & Div & TPR & PPL & Div \\
\hline
0.5 & 11.0\% & 10.03 & 7.02 & 42.0\% & 9.9 & 6.86 & 2.0\% & 9.33 & 7.9 & 29.0\% & 9.27 & 7.11 & 63.0\% & 8.92 & 7.09 \\
1.0 & 0.0\% & 9.57 & 6.72 & 5.0\% & 9.44 & 6.73 & 0.0\% & 10.4 & 8.64 & 6.0\% & 9.34 & 6.84 & 27.0\% & 9.19 & 6.84 \\
2.0 & 0.0\% & 9.35 & 6.65 & 0.0\% & 9.38 & 6.58 & 0.0\% & 11.16 & 8.26 & 1.0\% & 9.03 & 6.71 & 9.0\% & 8.89 & 6.59 \\
3.0 & 0.0\% & 9.45 & 6.46 & 1.0\% & 9.25 & 6.43 & 0.0\% & 11.33 & 8.61 & 0.0\% & 9.32 & 6.82 & 1.0\% & 9.05 & 6.65 \\
\midrule
\midrule
$\alpha$ & \multicolumn{3}{c}{X-SIR} & \multicolumn{3}{c}{UPV} & \multicolumn{3}{c}{Gumbel} & \multicolumn{3}{c}{EWD} & \multicolumn{3}{c}{SWEET} \\
\cmidrule(lr){2-4} \cmidrule(lr){5-7} \cmidrule(lr){8-10} \cmidrule(lr){11-13} \cmidrule(lr){14-16}
    & TPR & PPL & Div & TPR & PPL & Div & TPR & PPL & Div & TPR & PPL & Div & TPR & PPL & Div \\
\hline
0.5 & 28.0\% & 9.47 & 6.94 & 42.0\% & 10.01 & 7.14 & 80.0\% & 13.73 & 7.54 & 0.0\% & 9.76 & 7.01 & 6.0\% & 9.66 & 7.13 \\
1.0 & 9.0\% & 9.47 & 6.75 & 20.0\% & 10.01 & 6.89 & 9.0\% & 19.25 & 8.3 & 0.0\% & 9.93 & 6.78 & 0.0\% & 9.59 & 6.72 \\
2.0 & 6.0\% & 9.45 & 6.46 & 4.0\% & 9.28 & 6.59 & 0.0\% & 25.39 & 9.04 & 0.0\% & 9.63 & 6.58 & 0.0\% & 9.29 & 6.45 \\
3.0 & 0.0\% & 9.12 & 6.41 & 1.0\% & 9.85 & 6.57 & 0.0\% & 25.77 & 9.5 & 0.0\% & 9.43 & 6.68 & 0.0\% & 9.33 & 6.53 \\
\bottomrule
\end{tabular}
\end{table*}

%% file: data/tab_impact_k_llama_rotated.tex
\begin{table*}[ht!]
    \centering
    \caption{Effect of \( K \) on Smoothing Attack Performance (Llama3-8B). Evaluation of the smoothing attack's effectiveness against different watermarking algorithms on the Llama3-8B model, varying the number of top-\( K \) tokens accessible to the attacker.}
    \label{tab:impact_of_k_llama}
    \begin{tabular}{lccc|ccc|ccc|ccc}
    \toprule
    K & \multicolumn{3}{c}{KGW} & \multicolumn{3}{c}{Unigram} & \multicolumn{3}{c}{SynthID} & \multicolumn{3}{c}{DIP}  \\
    \cmidrule(lr){2-4} \cmidrule(lr){5-7} \cmidrule(lr){8-10} \cmidrule(lr){11-13} 
        & TPR & PPL & Div & TPR & PPL & Div & TPR & PPL & Div & TPR & PPL & Div\\
    \hline
    1  & 6\% & 2.37 & 4.67 & 19\% & 2.41 & 4.67 & 0\%  & 3.6  & 6.86 & 2\%  & 2.53 & 4.84 \\
    3  & 1\% & 2.81 & 5.17 & 27\% & 2.8  & 5.2  & 0\%  & 3.42 & 6.87 & 4\%  & 2.91 & 5.47 \\
    5  & 3\% & 2.99 & 5.36 & 24\% & 2.92 & 5.31 & 0\%  & 3.41 & 6.89 & 1\%  & 2.97 & 5.55 \\
    7  & 2\% & 3.14 & 5.55 & 23\% & 3.03 & 5.43 & 0\%  & 3.41 & 6.86 & 4\%  & 3.1  & 5.78 \\
    10 & 2\% & 3.2  & 5.63 & 24\% & 3.1  & 5.44 & 0\%  & 3.4  & 6.86 & 6\%  & 3.17 & 5.67 \\
    \bottomrule
    \toprule
    K & \multicolumn{3}{c}{Unbiased} & \multicolumn{3}{c}{UPV} & \multicolumn{3}{c}{EWD} & \multicolumn{3}{c}{SWEET}  \\
    \cmidrule(lr){2-4} \cmidrule(lr){5-7} \cmidrule(lr){8-10} \cmidrule(lr){11-13} 
        & TPR & PPL & Div & TPR & PPL & Div & TPR & PPL & Div & TPR & PPL & Div\\
    \hline
    1  & 1\% & 2.5  & 4.8  & 1\% & 2.48 & 4.76 & 3\% & 2.43 & 4.68 & 3\% & 2.41 & 4.72 \\
    3  & 4\% & 2.9  & 5.44 & 1\% & 2.97 & 5.37 & 4\% & 2.94 & 5.33 & 3\% & 2.91 & 5.27 \\
    5  & 2\% & 2.95 & 5.53 & 0\% & 3.02 & 5.55 & 3\% & 3.06 & 5.48 & 4\% & 3.01 & 5.43 \\
    7  & 7\% & 3.14 & 5.72 & 1\% & 3.1  & 5.54 & 6\% & 3.09 & 5.43 & 5\% & 3.01 & 5.37 \\
    10 & 5\% & 3.17 & 5.75 & 1\% & 3.12 & 5.49 & 3\% & 3.13 & 5.38 & 4\% & 3.09 & 5.4  \\
    \bottomrule
    \end{tabular}
    \end{table*}

%% file: data/tab_impact_alpha_llama_rotated.tex
\begin{table*}[ht]
    \centering
    \caption{Effect of \( \alpha \) on Smoothing Attack Performance (Llama3-8B). Evaluation of the smoothing attack's effectiveness against different watermarking algorithms on the Llama3-8B model, varying the parameter \( \alpha \). A larger \( \alpha \) indicates greater reliance on the reference model's output, while a smaller \( \alpha \) means the attack text is more influenced by the watermarked model.}
    \label{tab:impact_of_alpha_llama}
    \resizebox{0.9\linewidth}{!}{%
    \begin{tabular}{lccc|ccc|ccc|ccc}
    \toprule
    $\alpha$ & \multicolumn{3}{c}{KGW} & \multicolumn{3}{c}{Unigram} & \multicolumn{3}{c}{SynthID} & \multicolumn{3}{c}{DIP} \\
    \cmidrule(lr){2-4} \cmidrule(lr){5-7} \cmidrule(lr){8-10} \cmidrule(lr){11-13}
        & TPR & PPL & Div & TPR & PPL & Div & TPR & PPL & Div & TPR & PPL & Div \\
    \hline
    0.5 & 13\% & 3.45 & 5.92 & 62\% & 3.4  & 5.77 & 0\%  & 3.78 & 6.88 & 35\% & 3.34 & 6.19 \\
    1.0 & 2\%  & 3.2  & 5.63 & 24\% & 3.1  & 5.44 & 0\%  & 3.4  & 6.86 & 6\%  & 3.17 & 5.67 \\
    2.0 & 0\%  & 3.05 & 5.28 & 12\% & 2.93 & 5.21 & 0\%  & 3.49 & 6.87 & 3\%  & 2.99 & 5.23 \\
    3.0 & 0\%  & 2.93 & 5.17 & 12\% & 2.99 & 5.26 & 0\%  & 3.52 & 6.83 & 1\%  & 2.96 & 5.16 \\
    \midrule
    \midrule
    $\alpha$ & \multicolumn{3}{c}{Unbiased} & \multicolumn{3}{c}{UPV} & \multicolumn{3}{c}{EWD} & \multicolumn{3}{c}{SWEET} \\
    \cmidrule(lr){2-4} \cmidrule(lr){5-7} \cmidrule(lr){8-10} \cmidrule(lr){11-13}
        & TPR & PPL & Div & TPR & PPL & Div & TPR & PPL & Div & TPR & PPL & Div \\
    \hline
    0.5 & 26\% & 3.37 & 6.09 & 10\% & 3.47 & 6.08 & 28\% & 3.44 & 5.84 & 44\% & 3.38 & 5.9  \\
    1.0 & 5\%  & 3.17 & 5.75 & 1\%  & 3.12 & 5.49 & 3\%  & 3.13 & 5.38 & 4\%  & 3.09 & 5.4  \\
    2.0 & 3\%  & 2.98 & 5.28 & 0\%  & 2.96 & 5.2  & 0\%  & 3.0  & 5.36 & 1\%  & 3.06 & 5.38 \\
    3.0 & 3\%  & 2.96 & 5.21 & 0\%  & 2.99 & 5.2  & 0\%  & 2.89 & 5.18 & 0\%  & 2.93 & 5.22 \\
    \bottomrule
    \end{tabular}}
    \end{table*}

%% file: data/tab_text_example_gumbel.tex
\begin{table*}[ht]
\caption{Text generated by watermarked model and smoothing attack (OPT-1.3B against Gumbel sampling).}
    \label{tab:gumbel_example}
    \centering
\begin{tabular}{|p{7cm}|p{7cm}|} %
\hline
\textbf{Watermarked} & \textbf{Smoothing Attack} \\ \hline
\raggedright
Cluster comprises IBM's Opteron-based eServer 325 server and systems management software and storage devices that can run Linux and Windows operating systems.

The data center will be built on top of existing IBM Power servers. The company will offer a variety of services, including cloud-based services, as well as a "plug and play" environment that will allow users to transition to other hardware.

The data center will be built on top of existing IBM Power servers. The company will offer a variety of services, including cloud-based services, as well as a "plug and play" environment that will allow users to transition to other hardware. (Image: IBM)

The data center will be built on top of existing IBM Power servers. The company will offer a variety of services, including cloud-based services, as well as a "plug and play" environment that will allow users to transition to other hardware.

The data center will be built on top of existing IBM Power servers. The company will offer a variety of services, including cloud-based services, as well as a "plug and play" environment that will allow users to transition to other hardware.  & \raggedright 
Cluster comprises IBM's Opteron-based eServer 325 server and systems management software and storage devices that can run Linux and Windows operating systems.

IBM will start selling customers a prototype of the cluster by July, according to Jim Bessen, the executive vice president of middleware architecture and services at IBM.

“The cluster is just the tip of the iceberg,” he said.

The cluster firm computing is encased in a step-by-step process starting with activities such as how to set up clinks that back up disks, store data, create back-up schemes for data, and at what resolutions to save the data.

“We are not done yet,” he said. As new kinds of programming software and other technologies come out, users need to act on it.

While IBM expects internal clusters with the capability to run multiple operating systems in fuse over the next year, this capability will be available only to Enterprise Software Group (ESG) customers.

ESG will not sell its cluster technology to anyone else, Bessen said. 
\end{tabular}
\end{table*}

%% file: data/tab_impact_model_size_opt_rotated.tex
\begin{table*}[ht]
\centering
\caption{Impact of Model Size on the Smoothing Attack (OPT). Performance of the smoothing attack across different watermarking algorithms and various sizes of OPT models. The perplexity (PPL) is computed with respect to the OPT-30B model, while the reference model is consistently the OPT-125M. The table reports True Positive Rate (TPR), Perplexity (PPL), and Diversity (Div.) for unwatermarked, watermarked, and smoothed settings.}
\setlength{\tabcolsep}{4pt}
\label{tab:impact_model_size}
\resizebox{\linewidth}{!}{%
\begin{tabular}{ll|ccc|ccc|ccc|ccc|ccc}
\toprule
\multirow{2}{*}{Size} & \multirow{2}{*}{Setting} & \multicolumn{3}{c}{KGW} & \multicolumn{3}{c}{Unigram} & \multicolumn{3}{c}{SynthID} & \multicolumn{3}{c}{DIP} & \multicolumn{3}{c}{Unbiased} \\
\cmidrule(lr){3-5} \cmidrule(lr){6-8} \cmidrule(lr){9-11} \cmidrule(lr){12-14} \cmidrule(lr){15-17} 
    & & TPR & PPL & Div. & TPR & PPL & Div. & TPR & PPL & Div. & TPR & PPL & Div.  & TPR & PPL & Div.\\
\midrule
\multirow{3}{*}{1.3B} 
& Unwatermarked & 0.0\% & 12.95 & 8.67 & 0.0\% & 12.95 & 8.67 & 0.0\% & 12.95 & 8.67 & 0.0\% & 12.95 & 8.67 & 0.0\% & 12.95 & 8.67 \\
& Watermarked   & 100.0\% & 15.94 & 8.09 & 99.0\% & 16.53 & 7.29 & 100.0\% & 7.7 & 7.41 & 100.0\% & 15.16 & 8.44 & 99.0\% & 15.14 & 8.29 \\
& Smoothing     & 4.0\% & 10.48 & 6.72 & 6.0\% & 10.37 & 6.83 & 1.0\% & 11.37 & 8.67 & 6.0\% & 10.03 & 7.03 & 4.0\% & 9.94 & 6.79 \\
\midrule
\multirow{3}{*}{2.7B} 
& Unwatermarked & 0.0\% & 11.75 & 8.36 & 0.0\% & 11.75 & 8.36 & 0.0\% & 11.75 & 8.36 & 0.0\% & 11.75 & 8.36 & 0.0\% & 11.75 & 8.36 \\
& Watermarked   & 100.0\% & 13.94 & 7.88 & 100.0\% & 14.31 & 7.41 & 99.0\% & 6.86 & 7.55 & 97.0\% & 13.86 & 8.61 & 97.0\% & 13.6 & 8.69 \\
& Smoothing     & 4.0\% & 10.35 & 6.77 & 4.0\% & 10.35 & 6.66 & 6.0\% & 9.84 & 8.0 & 13.0\% & 9.87 & 6.84 & 6.0\% & 9.85 & 6.88 \\
\midrule
\multirow{3}{*}{6.7B} 
& Unwatermarked & 0.0\% & 10.2 & 8.45 & 0.0\% & 10.2 & 8.45 & 0.0\% & 10.2 & 8.45 & 0.0\% & 10.2 & 8.45 & 0.0\% & 10.2 & 8.45 \\
& Watermarked   & 100.0\% & 13.16 & 8.06 & 100.0\% & 12.94 & 7.48 & 98.0\% & 6.21 & 7.48 & 98.0\% & 11.8 & 8.48 & 97.0\% & 11.79 & 8.59 \\
& Smoothing     & 4.0\% & 10.07 & 6.92 & 6.0\% & 10.54 & 6.68 & 3.0\% & 8.98 & 8.31 & 8.0\% & 9.78 & 6.86 & 8.0\% & 9.68 & 6.74 \\
\midrule
\multirow{3}{*}{13B} 
& Unwatermarked & 0.0\% & 10.14 & 8.39 & 0.0\% & 10.14 & 8.39 & 0.0\% & 10.14 & 8.39 & 0.0\% & 10.14 & 8.39 & 0.0\% & 10.14 & 8.39 \\
& Watermarked   & 100.0\% & 12.88 & 8.56 & 100.0\% & 12.44 & 7.39 & 100.0\% & 5.88 & 7.8 & 96.0\% & 11.67 & 9.34 & 93.0\% & 11.42 & 8.77 \\
& Smoothing     & 2.0\% & 10.24 & 6.82 & 5.0\% & 10.32 & 6.7 & 8.0\% & 8.07 & 7.8 & 8.0\% & 9.6 & 6.88 & 7.0\% & 9.37 & 6.77 \\
\midrule
\multirow{3}{*}{30B} 
& Unwatermarked & 0.0\% & 8.46 & 8.44 & 0.0\% & 8.46 & 8.44 & 0.0\% & 8.46 & 8.44 & 0.0\% & 8.46 & 8.44 & 0.0\% & 8.46 & 8.44 \\
& Watermarked   & 100.0\% & 10.23 & 8.34 & 100.0\% & 10.45 & 7.56 & 100.0\% & 5.27 & 7.72 & 94.0\% & 9.43 & 8.78 & 97.0\% & 9.89 & 9.08 \\
& Smoothing     & 0.0\% & 9.5 & 6.8 & 7.0\% & 10.15 & 6.75 & 5.0\% & 6.96 & 8.04 & 4.0\% & 9.34 & 6.89 & 4.0\% & 9.36 & 6.88 \\
\midrule
\midrule
\multirow{2}{*}{Size} & \multirow{2}{*}{Setting} & \multicolumn{3}{c}{X-SIR} & \multicolumn{3}{c}{UPV} & \multicolumn{3}{c}{Gumbel} & \multicolumn{3}{c}{EWD} & \multicolumn{3}{c}{SWEET} \\
\cmidrule(lr){3-5} \cmidrule(lr){6-8} \cmidrule(lr){9-11} \cmidrule(lr){12-14} \cmidrule(lr){15-17} 
    & & TPR & PPL & Div. & TPR & PPL & Div. & TPR & PPL & Div. & TPR & PPL & Div.  & TPR & PPL & Div.\\
\midrule
\multirow{3}{*}{1.3B} 
& Unwatermarked & 1.0\% & 12.95 & 8.67 & 0.0\% & 12.95 & 8.67 & 0.0\% & 12.95 & 8.67 & 0.0\% & 12.95 & 8.67 & 0.0\% & 12.95 & 8.67 \\
& Watermarked   & 94.0\% & 15.42 & 7.96 & 99.0\% & 12.79 & 8.22 & 98.0\% & 3.15  & 4.35 & 100.0\% & 16.88 & 7.92 & 100.0\% & 15.99 & 8.02 \\
& Smoothing     & 13.0\% & 10.3 & 6.72 & 20.0\% & 10.78 & 6.89 & 9.0\% & 20.94 & 8.30 & 1.0\% & 10.71 & 6.75 & 1.0\% & 10.54 & 6.81 \\
\midrule
\multirow{3}{*}{2.7B} 
& Unwatermarked & 3.0\% & 11.75 & 8.36 & 0.0\% & 11.75 & 8.36 & 0.0\% & 11.75 & 8.36 & 0.0\% & 11.75 & 8.36 & 0.0\% & 11.75 & 8.36 \\
& Watermarked   & 91.0\% & 14.07 & 8.25 & 99.0\% & 12.30 & 8.01 & 99.0\% & 2.96  & 4.38 & 100.0\% & 14.88 & 7.98 & 100.0\% & 14.07 & 8.32 \\
& Smoothing     & 10.0\% & 10.34 & 6.77 & 18.0\% & 10.56 & 6.90 & 10.0\% & 19.46 & 8.41 & 1.0\% & 10.43 & 6.86 & 3.0\% & 10.49 & 6.86 \\
\midrule
\multirow{3}{*}{6.7B} 
& Unwatermarked & 0.0\% & 10.2 & 8.45 & 0.0\% & 10.20 & 8.45 & 0.0\% & 10.20 & 8.45 & 0.0\% & 10.20 & 8.45 & 0.0\% & 10.20 & 8.45 \\
& Watermarked   & 91.0\% & 13.04 & 8.19 & 97.0\% & 10.92 & 7.75 & 100.0\% & 2.97  & 4.49 & 100.0\% & 13.42 & 8.69 & 100.0\% & 13.05 & 8.41 \\
& Smoothing     & 9.0\% & 10.01 & 6.7 & 8.0\% & 10.60 & 7.05 & 9.0\% & 14.85 & 8.62 & 0.0\% & 10.60 & 6.79 & 1.0\% & 10.07 & 6.89 \\
\midrule
\multirow{3}{*}{13B} 
& Unwatermarked & 0.0\% & 10.14 & 8.39 & 0.0\% & 10.14 & 8.39 & 0.0\% & 10.14 & 8.39 & 0.0\% & 10.14 & 8.39 & 0.0\% & 10.14 & 8.39 \\
& Watermarked   & 88.0\% & 12.29 & 8.05 & 99.0\% & 10.59 & 7.91 & 98.0\% & 2.96  & 4.63 & 100.0\% & 13.09 & 8.74 & 100.0\% & 12.32 & 8.35 \\
& Smoothing     & 11.0\% & 9.84 & 6.79 & 12.0\% & 10.84 & 6.88 & 12.0\% & 15.06 & 8.27 & 0.0\% & 10.16 & 6.73 & 2.0\% & 10.15 & 6.74 \\
\midrule
\multirow{3}{*}{30B} 
& Unwatermarked & 0.0\% & 8.46 & 8.44 & 0.0\% & 8.46 & 8.44 & 0.0\% & 8.46 & 8.44 & 0.0\% & 8.46 & 8.44 & 0.0\% & 8.46 & 8.44 \\
& Watermarked   & 91.0\% & 10.43 & 8.43 & 97.0\% & 8.59 & 8.13 & 97.0\% & 2.89  & 4.79 & 100.0\% & 10.75 & 8.54 & 100.0\% & 9.98 & 8.25 \\
& Smoothing     & 16.0\% & 9.65 & 6.74 & 17.0\% & 10.06 & 7.11 & 9.0\% & 11.92 & 8.39 & 2.0\% & 10.02 & 6.99 & 2.0\% & 9.55 & 6.84 \\
\bottomrule
\end{tabular}}
\end{table*}

%% file: sec/appendix_analysis.tex
\section{Analysis}
\label{appendix:analytical_analysis}
\subsection{Contribution depends on the model confidence}
\label{appendix:contribution_vs_pg}
\label{appendix:contritbuion_vs_l2}

We first demonstrate that the contribution of each token to the detection score is influenced by the confidence score of the unwatermarked model, as measured by its probability distribution.

\subsubsection{Case Study: Green-Red List Watermark}

Suppose that $l_t$ is the logit vector for predicting the $t$-th token from the unwatermarked model, and $\mathcal{G}_t$ is the green list used by the watermarked model at position $t$, with size $\gamma |\mathcal{V}|$. Given the watermark shift $\delta$, the probabilities assigned by the unwatermarked and watermarked models are expressed as:

\begin{align*}
P_t(v) \;&=\; 
\frac{\exp\!\bigl(l_t(v)\bigr)}{\sum_{v' \,\in\, \mathcal{V}} \exp\!\bigl(l_t(v')\bigr)}.\\
\widetilde{P}_t(v) 
\;&=\;
\frac{\exp\!\bigl(l_t(v) + \delta \cdot \mathbf{1}_{\{v \in \mathcal{G}_t\}}\bigr)}%
     {\sum_{v' \,\in\, \mathcal{V}} \exp\!\Bigl(l_t(v') + \delta \cdot \mathbf{1}_{\{v' \in \mathcal{G}_t\}}\Bigr)}.
 \end{align*}

Rewriting $\widetilde{P}_t(v)$, we observe:
\[
\widetilde{P}_t(v) 
\;=\;
P_t(v) \times
\frac{\exp\!\bigl(\delta\,\mathbf{1}_{\{v \in \mathcal{G}_t\}}\bigr)}%
     {\sum_{v' \,\in\, \mathcal{V}} P_t(v')\, \exp\!\bigl(\delta\,\mathbf{1}_{\{v' \in \mathcal{G}_t\}}\bigr)}.
\]

Define the normalization factor:
\[
Z_\delta 
\;=\; \frac{\sum_{v' \in \mathcal{V}} \exp\!\bigl(l_t(v') + \delta \,\mathbf{1}_{\{v' \in \mathcal{G}_t\}}\bigr)}{\sum_{v' \in \mathcal{V}} \exp\!\bigl(l_t(v') \bigr)}
\;=\; 
\sum_{v' \,\in\, \mathcal{V}} P_t(v') \,
\exp\!\bigl(\delta \,\mathbf{1}_{\{v' \in \mathcal{G}_t\}}\bigr).
\]

Then:
\[
\widetilde{P}_t(v)
\;=\;
\begin{cases}
\displaystyle
\frac{e^\delta}{Z_\delta}\;P_t(v), 
& v \in \mathcal{G}_t, \\[6pt]
\displaystyle
\frac{1}{Z_\delta}\;P_t(v), 
& v \notin \mathcal{G}_t.
\end{cases}
\]

The expected fraction of tokens belonging to the green list under the unwatermarked model is given by:

\[
    \mathbb{E}_{v \sim P_t}[\mathbf{1}(v \in \mathcal{G}_t)] 
    \;=\; \sum_{v \in \mathcal{G}_t} P_t(v) \;=\; P_{\mathcal{G}_t},
\]
where $P_{\mathcal{G}_t}$ represents the probability mass assigned to green tokens in the unwatermarked model.

Similarly, the expected fraction of green tokens in the watermarked model is:

\begin{align}
    \mathbb{E}_{v \sim \widetilde{P}_t}[\mathbf{1}(v \in \mathcal{G}_t)] = \sum_{v \in \mathcal{G}_t} \widetilde{P}_t(v) 
    & \;=\; \frac{e^\delta}{Z_\delta}\; P_{\mathcal{G}_t}.
\end{align}

Since $Z_\delta = (e^\delta -1)P_{\mathcal{G}_t} + 1$, the difference in green token probabilities (i.e., the detection contribution at token position $t$) is:

\begin{align}
    S_t = \mathbb{E}_{v \sim \widetilde{P}_t}[\mathbf{1}(v \in \mathcal{G}_t)] - \mathbb{E}_{v \sim P_t}[\mathbf{1}(v \in \mathcal{G}_t)]
    = \frac{-(e^\delta - 1) P_{\mathcal{G}_t} + (e^\delta - 1)}{(e^\delta - 1) + \frac{1}{P_{\mathcal{G}_t}}}.
\end{align}

In other words, the token-level detection contribution $S_t$ is a function of the probability mass $P_{\mathcal{G}_t}$ assigned to green tokens by the unwatermarked model.

\subsubsection{Case Study: Tournament Sampling Watermark}

In the Tournament Sampling watermark, when generating the $t$-th token, the algorithm assigns scores to each token using $m$ independent watermarking functions $g^{(1)}, ..., g^{(m)}$. These scores depend on a random seed generated based on the recent context and a secret watermarking key. The token selection follows a multi-round elimination process, where $2^m$ tokens are first sampled from $P_t(\cdot)$, then compete in $m$ rounds to determine the final output.

Despite the complex sampling mechanism, the probability of each token in the modified distribution $\widetilde{P}_t$ is adjusted by a factor dependent on its assigned $g$ value. Specifically, for any token $v$:

\begin{align}
\widetilde{P}_t(v) = 
\begin{cases} 
P_t(v) \cdot (1 - P_{\mathcal{G}_t})  & \text{if } g(v) = 0, \\
P_t(v) \cdot (2 - P_{\mathcal{G}_t})  & \text{if } g(v) = 1.
\end{cases}
\end{align}

During watermark detection, the detector computes the average $g$ value across all tournament layers, i.e., $\frac{1}{m} \sum_{l=1}^{m} g^{(l)}(v)$, as the watermark score for the token.

\noindent\textit{Single Tournament Layer ($m=1$).}  
Consider the simplest case where $m=1$, meaning only one tournament round is used. Let $\mathcal{G}_t$ denote the set of tokens where $g^{(1)}(v) = 1$. The probability modification simplifies to:

\begin{align}
\widetilde{P}_t(v) = 
\begin{cases} 
P_t(v) \cdot (1 - P_{\mathcal{G}_t})  & \text{if } v \notin \mathcal{G}_t, \\
P_t(v) \cdot (2 - P_{\mathcal{G}_t}) & \text{if } v \in \mathcal{G}_t.
\end{cases}
\end{align}

The expected $g$ value for tokens sampled from $\widetilde{P}_t$ is $(2 - P_{\mathcal{G}_t}) \cdot P_{\mathcal{G}_t}$, while the expectation under $P_t$ is simply $P_{\mathcal{G}_t}$. Thus, the detection contribution $S_t$ is:

\begin{align}
	S_t = (1 - P_{\mathcal{G}_t}) \cdot P_{\mathcal{G}_t}.
\end{align} 

This mirrors the Green-Red List watermark, showing that the detection contribution per token is fundamentally tied to $P_{\mathcal{G}_t}$.

Thus far, we have established that the contribution of each token to the detection score is correlated with the expected watermark score under the unwatermarked model. We now analyze what affects the watermark score of the unwatermarked model.

Let $P_t = (p_1, p_2, \dots, p_d)$ be the probability vector from the unwatermarked model at token position $t$, where $p_i \in [0,1]$ and $\sum_{i=1}^d p_i = 1$. Typically, $d = |\mathcal{V}|$ is large. We randomly select a subset $\mathcal{G}_t \subset \{1,\ldots,d\}$ of indices of size $\gamma |\mathcal{V}|$. Define the random variable:

\[
  P_{\mathcal{G}_t} \;=\; \sum_{i \in \mathcal{G}_t} p_i.
\]

We analyze how $P_{\mathcal{G}_t}$ is distributed over all possible assignments of $\mathcal{G}_t$. Define the indicator variable $X_i$ as follows:

\[
   X_i \;=\;
   \begin{cases}
     1, & \text{if } i \in \mathcal{G}_t,\\
     0, & \text{otherwise.}
   \end{cases}
\]

Since each token is independently assigned to $\mathcal{G}_t$ with probability $\gamma$, we have:

\[
\mathbb{E}[X_i] = \gamma, \quad \text{and} \quad \mathrm{Var}(X_i) = \gamma(1-\gamma).
\]

For different token indices $i \neq j$, the covariance between their assignments is:

\[
   \mathrm{Cov}(X_i, X_j) = \mathbb{E}[X_i X_j] - \mathbb{E}[X_i]\mathbb{E}[X_j].
\]

For Poisson sampling (i.e., assigning each token to $\mathcal{G}_t$ independently with probability $\gamma$), the covariance is zero. However, under a fixed-size sampling setup (i.e., selecting exactly $\gamma |\mathcal{V}|$ tokens), we have:

\[
   \mathrm{Cov}(X_i, X_j) = \frac{\gamma |\mathcal{V}|}{d} \cdot \frac{\gamma |\mathcal{V}| - 1}{d - 1} - \gamma^2
   = -\frac{\gamma(1-\gamma)}{|\mathcal{V}| - 1}.
\]

Expressing $P_{\mathcal{G}_t}$ in terms of $X_i$, we obtain:

\[
    P_{\mathcal{G}_t} \;=\; \sum_{i=1}^d X_i\,p_i.
\]

\paragraph{Expectation and Variance of $P_{\mathcal{G}_t}$.}
The expectation is:

\[
  \mathbb{E}[ P_{\mathcal{G}_t}] 
  \;=\; 
  \sum_{i=1}^d \mathbb{E}[X_i]\,p_i 
  \;=\; 
  \gamma \sum_{i=1}^d p_i 
  \;=\; 
  \gamma.
\]

The variance is:

\[
\mathrm{Var}( P_{\mathcal{G}_t}) = \sum_{i=1}^d p_i^2 \mathrm{Var}(X_i) + \sum_{i \neq j} p_i p_j \mathrm{Cov}(X_i, X_j).
\]

Substituting $\mathrm{Var}(X_i) = \gamma(1-\gamma)$ and $\mathrm{Cov}(X_i, X_j) = -\frac{\gamma(1-\gamma)}{|\mathcal{V}| - 1}$:

\[
\mathrm{Var}(P_{\mathcal{G}_t}) = \gamma(1-\gamma) \sum_{i=1}^d p_i^2 - \frac{\gamma(1-\gamma)}{|\mathcal{V}| - 1} \sum_{i \neq j} p_i p_j.
\]

For the first term,

\[
\gamma(1-\gamma) \sum_{i=1}^d p_i^2 = \gamma(1-\gamma) \sigma^2,
\]

where $\sigma^2 = \sum_{i=1}^d p_i^2$ represents the squared $\ell_2$ norm of the probability vector.

For the second term, using the identity:

\[
   \sum_{i \neq j} p_i p_j = \left(\sum_{i=1}^d p_i\right)^2 - \sum_{i=1}^d p_i^2 = 1 - \sigma^2,
\]

and we obtain:

\[
\frac{\gamma(1-\gamma)}{|\mathcal{V}| - 1} \sum_{i \neq j} p_i p_j = \frac{\gamma(1-\gamma)}{|\mathcal{V}| - 1} (1 - \sigma^2).
\]

For large $|\mathcal{V}|$, the correction term $\frac{\gamma(1-\gamma)}{|\mathcal{V}| - 1} (1 - \sigma^2)$ becomes negligible, and we approximate:

\[
\mathrm{Var}(P_{\mathcal{G}_t}) \approx \gamma(1-\gamma) \sigma^2.
\]

\paragraph{Interpretation.}
This analysis shows that $P_{\mathcal{G}_t}$ depends on the probability mass distribution.

High-Uncertainty Case (Uniform Distribution):  
  If $p_i = \frac{1}{|\mathcal{V}|}$ for all $i$, then

  \[
  \sigma^2 = \sum_{i=1}^{|\mathcal{V}|} \frac{1}{|\mathcal{V}|^2} = \frac{1}{|\mathcal{V}|}.
  \]

  For large $|\mathcal{V}|$, $\sigma^2$ is small, meaning that the distribution of $P_{\mathcal{G}_t}$ concentrates tightly around $\gamma$ with small variance. This corresponds to a scenario where the model has high uncertainty, spreading probability mass nearly uniformly over all tokens.

Low-Uncertainty Case (Dominant Tokens): 
  In practice, language models often assign high probability mass to a small number of dominant tokens. Suppose $p_j \geq 0.8$ for some token $j$, then:

  \[
  \sigma^2 \geq p_j^2 = 0.64.
  \]

  In this case, $\sigma^2$ is much larger than $1/|\mathcal{V}|$ (which is on the order of $10^{-5}$ for large models). Consequently, $P_{\mathcal{G}_t}$ exhibits a bimodal distribution: it is either close to $0$ or close to $1$, depending on whether the dominant tokens are in $\mathcal{G}_t$. The probability of $P_{\mathcal{G}_t} \approx \gamma$ is nearly zero.

Thus, when the model is confident in its predictions (low uncertainty), the variance of $P_{\mathcal{G}_t}$ is large, leading to a higher variance in the watermark score. Conversely, when the model is uncertain, the watermark score is more stable and centered around $\gamma$.

\paragraph{Connection to Watermark Detection.}
Since the contribution to the detection score $S_t$ depends on $P_{\mathcal{G}_t}$ (Eq.~\eqref{eq:contribution-green}), its variance is governed by $\mathrm{Var}(P_{\mathcal{G}_t})$. This means that tokens generated with high confidence contribute more variability to the detection score, whereas tokens generated under uncertainty contribute less variability.

\subsection{Estimating the confidence score}\label{appendix:estimate_l2}

Our goal is to estimate the squared $\ell_2$ norm of the probability distribution $\|P_t\|^2$, which serves as a confidence measure for the unwatermarked model, using only access to the watermarked model $\widetilde{P}_t$. This estimation is critical for adaptive attacks and for understanding how watermarking affects text quality.

\paragraph{Setup.} 
We consider the Green-Red List watermarking scheme, where the probability distribution $\widetilde{P}_t$ is obtained by modifying $P_t$ as:

\[
  \widetilde{P}_t(v) 
  \;=\;
  \frac{e^{\delta \mathbf{1}_{\{v \in \mathcal{G}_t\}}}}{Z_\delta} P_t(v),
\]

where the normalization factor $Z_\delta$ is defined as:

\[
  Z_\delta = (1 - P_{\mathcal{G}_t}) + e^\delta P_{\mathcal{G}_t}.
\]

We aim to construct an estimator $\widehat{U}$ for the confidence measure:

\[
  \|P_t\|^2 = \sum_{v \in \mathcal{V}} P_t(v)^2.
\]

\paragraph{Expected Squared Norm of the Watermarked Model.} 
Since each probability mass in $P_t$ is scaled by either $e^\delta / Z_\delta$ (if in $\mathcal{G}_t$) or $1/Z_\delta$ (if not in $\mathcal{G}_t$), we have:

\[
  \mathbb{E}[\widetilde{P}_t(v)^2] 
  \;=\; (1 - \gamma) \frac{1}{Z_\delta^2} P_t(v)^2 + \gamma \frac{e^{2\delta}}{Z_\delta^2} P_t(v)^2.
\]

Summing over all tokens in $\mathcal{V}$, we obtain:

\[
  \mathbb{E}[\|\widetilde{P}_t\|^2] 
  \;=\; \frac{(1 - \gamma) + \gamma e^{2\delta}}{Z_\delta^2} \|P_t\|^2.
\]

\paragraph{Unbiased Estimator.} 
Rearranging the above expression, we define an unbiased estimator:

\[
  \widehat{U} = \frac{Z_\delta^2}{(1 - \gamma) + \gamma e^{2\delta}} \|\widetilde{P}_t\|^2.
\]

Taking expectation, we confirm:

\[
  \mathbb{E}[\widehat{U}] = \|P_t\|^2.
\]

\paragraph{Practical Approximation.} 
Since $Z_\delta$ depends on $P_{\mathcal{G}_t}$, which is unknown to an adversary, we approximate it using $\gamma$:

\[
  Z_\delta \approx (1 - \gamma) + \gamma e^\delta.
\]

Thus, the practical estimator becomes:

\[
  \widetilde{U} = \frac{[(1 - \gamma) + \gamma e^\delta]^2}{(1 - \gamma) + \gamma e^{2\delta}} \|\widetilde{P}_t\|^2.
\]

This provides a computationally efficient way to estimate $\|P_t\|^2$ using only $\widetilde{P}_t$, making it useful for designing attacks.

\subsection{Estimating the confidence score using top-$K$ probabilities}\label{appendix:estimate_l2_using_topk}

While we have established the connection between the squared $\ell_2$ norm $\|P_t\|^2$ of the probability distribution and its contribution to the watermark detection score, direct access to this quantity is often unavailable, even for the watermarked model. In this section, we show how to estimate $\|P_t\|^2$ using only limited access to the model’s top-$K$ probabilities.

Suppose we only have access to the {top-$K$} probabilities:

\[
p_1 \ge p_2 \ge \dots \ge p_K,
\]

where the remaining probabilities $p_{K+1}, \dots, p_{|\mathcal{V}|}$ are unknown. Define the {remaining probability mass} of the tail as:

\[
R = 1 - \sum_{i=1}^{K} p_i.
\]

Our goal is to estimate the squared $\ell_2$ norm:

\[
\|P_t\|^2 = \sum_{i=1}^{|\mathcal{V}|} p_i^2,
\]

given only $p_1, \dots, p_K$ and $R$.

We bound $\|P_t\|^2$ by considering two extreme ways in which the unknown tail probabilities could be distributed:

\begin{enumerate}
  \item {Uniform Tail:} The remaining probability mass $R$ is evenly distributed across the unknown tokens, minimizing the sum of squares.
  \item {Concentrated Tail:} The entire probability mass $R$ is assigned to a single token, maximizing the sum of squares.
\end{enumerate}

\paragraph{Uniform Tail (Lower Bound)}
If the tail probability mass $R$ is \emph{uniformly} spread among the remaining $|\mathcal{V}| - K$ tokens, then each unknown probability is $\frac{R}{|\mathcal{V}| - K}$. The squared sum of the tail probabilities is then:

\[
\sum_{i=K+1}^{|\mathcal{V}|} p_i^2 
\;=\;
(|\mathcal{V}| - K) \left(\frac{R}{|\mathcal{V}| - K}\right)^2 
\;=\; 
\frac{R^2}{|\mathcal{V}| - K}.
\]

Since distributing the mass uniformly minimizes the squared sum (due to convexity), this scenario provides a {lower bound} for $\|P_t\|^2$:

\[
\|P_t\|^2
\;\ge\;
\sum_{i=1}^{K} p_i^2 + \frac{R^2}{|\mathcal{V}| - K}.
\]

\paragraph{Concentrated Tail (Upper Bound)}
At the other extreme, if the entire remaining probability mass $R$ is assigned to a single token, then the squared sum of the tail probabilities is simply:

\[
\sum_{i=K+1}^{|\mathcal{V}|} p_i^2 = R^2.
\]

Since concentrating all probability mass in one entry maximizes the sum of squares, this provides an {upper bound} for $\|P_t\|^2$:

\[
\|P_t\|^2
\;\le\;
\sum_{i=1}^{K} p_i^2 + R^2.
\]

Combining both bounds, we obtain:

\[
\sum_{i=1}^{K} p_i^2 
+ \frac{R^2}{|\mathcal{V}| - K}
\;\;\le\;\;
\|P_t\|^2
\;\;\le\;\;
\sum_{i=1}^{K} p_i^2 
+ R^2,
\]

where $R = 1 - \sum_{i=1}^{K} p_i$.

\paragraph{Practical Approximation.}
A commonly used practical heuristic is to assume that the remaining probability mass $R$ follows a uniform distribution across the unknown probabilities. Under this assumption, we approximate:

\[
\|P_t\|^2 
\;\approx\;
\sum_{i=1}^{K} p_i^2 
+ \frac{R^2}{|\mathcal{V}| - K}.
\]

This estimate tends to be slightly lower than the true value, since in reality, the tail probabilities are rarely perfectly uniform—some tokens may have slightly higher probabilities than others. However, in the case of language modeling, probability distributions often exhibit a ``long tail'' where the remaining probability mass is spread across many small values. In such cases, the uniform assumption serves as a reasonable first-order approximation.

%% file: sec/appendix_limitations.tex
\section{Possible Defenses to Smoothing Attack}
\label{appendix:possible_defense}
Our attack exploits the correlation between a token's contribution to the watermark detection score and the confidence level of the unwatermarked model in predicting that token. One possible defense against this attack is to restrict access to confidence-related information, such as returning only the most probable token without revealing its probability. Note that, if the probability of the most likely token is available, our attack remains effective.

However, such a defense is challenging to enforce in practice. Many existing LLM services provide \textit{top-$K$ probabilities} (e.g., OpenAI's API returns probabilities for the top 20 tokens), which is already sufficient to approximate model confidence and execute our attack. Moreover, service providers often release these probabilities to enhance transparency and build trust by providing insights into the model’s reasoning, addressing concerns about the opacity of AI systems \cite{euaiact, oecdaiprinciples}. 

Access to probability distributions is also essential for debugging and evaluating model performance, as it allows developers to identify biases, diagnose overconfidence, and improve reliability \cite{nistframework}. Probabilities support explainable AI (XAI) by revealing model uncertainty, enabling users to interpret predictions and explore alternative suggestions \cite{openairesearch}. From an ethical standpoint, making probability distributions available facilitates bias auditing and aligns with broader efforts to promote fairness and accountability in AI \cite{oecdaiprinciples}. Additionally, probability information empowers developers and end users by enabling advanced decision-making strategies, such as re-ranking, rejection sampling, and beam search \cite{openaiapidocs}. Furthermore, it helps mitigate risks associated with model overconfidence and hallucinations, which is particularly crucial in high-stakes domains such as healthcare and law \cite{nistframework}.

Given the practical difficulties in restricting access to confidence-related information, our findings suggest that existing watermarking techniques may be vulnerable when model confidence can be estimated. This highlights the need for developing watermarking schemes that remain effective even in scenarios where adversaries have partial access to confidence estimates. Future research should explore watermarking methods that explicitly account for the model's confidence and ensure robustness against adversarial attacks that exploit confidence information.

%% file: main.bbl
\begin{thebibliography}{55}
\providecommand{\natexlab}[1]{#1}
\providecommand{\url}[1]{#1}
\csname url@samestyle\endcsname
\providecommand{\newblock}{\relax}
\providecommand{\bibinfo}[2]{#2}
\providecommand{\BIBentrySTDinterwordspacing}{\spaceskip=0pt\relax}
\providecommand{\BIBentryALTinterwordstretchfactor}{4}
\providecommand{\BIBentryALTinterwordspacing}{\spaceskip=\fontdimen2\font plus
\BIBentryALTinterwordstretchfactor\fontdimen3\font minus
  \fontdimen4\font\relax}
\providecommand{\BIBforeignlanguage}[2]{{%
\expandafter\ifx\csname l@#1\endcsname\relax
\typeout{** WARNING: IEEEtranN.bst: No hyphenation pattern has been}%
\typeout{** loaded for the language `#1'. Using the pattern for}%
\typeout{** the default language instead.}%
\else
\language=\csname l@#1\endcsname
\fi
#2}}
\providecommand{\BIBdecl}{\relax}
\BIBdecl

\bibitem[Aaronson(2023)]{aaronson2023watermarking}
S.~Aaronson, ``Simons institute talk on watermarking of large language
  models,''
  \url{https://simons.berkeley.edu/talks/scott-aaronson-ut-austin-openai-2023-08-17},
  2023.

\bibitem[Christ et~al.(2023)Christ, Gunn, and Zamir]{christ2023undetectable}
M.~Christ, S.~Gunn, and O.~Zamir, ``Undetectable watermarks for language
  models,'' \emph{arXiv preprint arXiv:2306.09194}, 2023.

\bibitem[Huang et~al.(2023)Huang, Zhu, Zhu, Lee, Jiao, and
  Jordan]{huang2023towards}
B.~Huang, B.~Zhu, H.~Zhu, J.~D. Lee, J.~Jiao, and M.~I. Jordan, ``Towards
  optimal statistical watermarking,'' \emph{arXiv preprint arXiv:2312.07930},
  2023.

\bibitem[Li et~al.(2024)Li, Ruan, Wang, Long, and Su]{li2024statistical}
X.~Li, F.~Ruan, H.~Wang, Q.~Long, and W.~J. Su, ``A statistical framework of
  watermarks for large language models: Pivot, detection efficiency and optimal
  rules,'' \emph{arXiv preprint arXiv:2404.01245}, 2024.

\bibitem[Dathathri et~al.(2024)Dathathri, See, Ghaisas, Huang, McAdam, Welbl,
  Bachani, Kaskasoli, Stanforth, Matejovicova, et~al.]{dathathri2024scalable}
S.~Dathathri, A.~See, S.~Ghaisas, P.-S. Huang, R.~McAdam, J.~Welbl, V.~Bachani,
  A.~Kaskasoli, R.~Stanforth, T.~Matejovicova \emph{et~al.}, ``Scalable
  watermarking for identifying large language model outputs,'' \emph{Nature},
  vol. 634, no. 8035, pp. 818--823, 2024.

\bibitem[Zhang et~al.(2022)Zhang, Roller, Goyal, Artetxe, Chen, Chen, Dewan,
  Diab, Li, Lin, et~al.]{zhang2022opt}
S.~Zhang, S.~Roller, N.~Goyal, M.~Artetxe, M.~Chen, S.~Chen, C.~Dewan, M.~Diab,
  X.~Li, X.~V. Lin \emph{et~al.}, ``Opt: Open pre-trained transformer language
  models,'' \emph{arXiv preprint arXiv:2205.01068}, 2022.

\bibitem[Dubey et~al.(2024)Dubey, Jauhri, Pandey, Kadian, Al-Dahle, Letman,
  Mathur, Schelten, Yang, Fan, et~al.]{dubey2024llama}
A.~Dubey, A.~Jauhri, A.~Pandey, A.~Kadian, A.~Al-Dahle, A.~Letman, A.~Mathur,
  A.~Schelten, A.~Yang, A.~Fan \emph{et~al.}, ``The llama 3 herd of models,''
  \emph{arXiv preprint arXiv:2407.21783}, 2024.

\bibitem[Chu et~al.(2024)Chu, Xu, Yang, Wei, Wei, Guo, Leng, Lv, He, Lin,
  et~al.]{chu2024qwen2}
Y.~Chu, J.~Xu, Q.~Yang, H.~Wei, X.~Wei, Z.~Guo, Y.~Leng, Y.~Lv, J.~He, J.~Lin
  \emph{et~al.}, ``Qwen2-audio technical report,'' \emph{arXiv preprint
  arXiv:2407.10759}, 2024.

\bibitem[Fan et~al.(2018)Fan, Lewis, and Dauphin]{fan2018hierarchical}
A.~Fan, M.~Lewis, and Y.~Dauphin, ``Hierarchical neural story generation,'' in
  \emph{Proceedings of the 56th Annual Meeting of the Association for
  Computational Linguistics (Volume 1: Long Papers)}, 2018, pp. 889--898.

\bibitem[Holtzman et~al.(2018)Holtzman, Buys, Forbes, Bosselut, Golub, and
  Choi]{holtzman2018learning}
A.~Holtzman, J.~Buys, M.~Forbes, A.~Bosselut, D.~Golub, and Y.~Choi, ``Learning
  to write with cooperative discriminators,'' in \emph{Proceedings of the 56th
  Annual Meeting of the Association for Computational Linguistics (Volume 1:
  Long Papers)}, 2018, pp. 1638--1649.

\bibitem[Holtzman et~al.(2019)Holtzman, Buys, Du, Forbes, and
  Choi]{holtzman2019curious}
A.~Holtzman, J.~Buys, L.~Du, M.~Forbes, and Y.~Choi, ``The curious case of
  neural text degeneration,'' in \emph{International Conference on Learning
  Representations}, 2019.

\bibitem[Kirchenbauer et~al.(2023{\natexlab{a}})Kirchenbauer, Geiping, Wen,
  Katz, Miers, and Goldstein]{kirchenbauer23a}
\BIBentryALTinterwordspacing
J.~Kirchenbauer, J.~Geiping, Y.~Wen, J.~Katz, I.~Miers, and T.~Goldstein, ``A
  watermark for large language models,'' in \emph{Proceedings of the 40th
  International Conference on Machine Learning}, ser. Proceedings of Machine
  Learning Research, A.~Krause, E.~Brunskill, K.~Cho, B.~Engelhardt, S.~Sabato,
  and J.~Scarlett, Eds., vol. 202.\hskip 1em plus 0.5em minus 0.4em\relax PMLR,
  23--29 Jul 2023, pp. 17\,061--17\,084. [Online]. Available:
  \url{https://proceedings.mlr.press/v202/kirchenbauer23a.html}
\BIBentrySTDinterwordspacing

\bibitem[Kuditipudi et~al.(2023)Kuditipudi, Thickstun, Hashimoto, and
  Liang]{kuditipudi2023robust}
R.~Kuditipudi, J.~Thickstun, T.~Hashimoto, and P.~Liang, ``Robust
  distortion-free watermarks for language models,'' \emph{arXiv preprint
  arXiv:2307.15593}, 2023.

\bibitem[Gumbel(1954)]{gumbel1954statistical}
E.~J. Gumbel, ``Statistical theory of extreme valuse and some practical
  applications,'' \emph{Nat. Bur. Standards Appl. Math. Ser. 33}, 1954.

\bibitem[Kirchenbauer et~al.(2023{\natexlab{b}})Kirchenbauer, Geiping, Wen,
  Shu, Saifullah, Kong, Fernando, Saha, Goldblum, and
  Goldstein]{kirchenbauer2023reliability}
J.~Kirchenbauer, J.~Geiping, Y.~Wen, M.~Shu, K.~Saifullah, K.~Kong,
  K.~Fernando, A.~Saha, M.~Goldblum, and T.~Goldstein, ``On the reliability of
  watermarks for large language models,'' \emph{arXiv preprint
  arXiv:2306.04634}, 2023.

\bibitem[Lee et~al.(2023)Lee, Hong, Ahn, Hong, Lee, Yun, Shin, and
  Kim]{lee2023wrote}
T.~Lee, S.~Hong, J.~Ahn, I.~Hong, H.~Lee, S.~Yun, J.~Shin, and G.~Kim, ``Who
  wrote this code? watermarking for code generation,'' \emph{arXiv preprint
  arXiv:2305.15060}, 2023.

\bibitem[Liu et~al.(2023)Liu, Pan, Hu, Li, Wen, King, and
  Philip]{liu2023unforgeable}
A.~Liu, L.~Pan, X.~Hu, S.~Li, L.~Wen, I.~King, and S.~Y. Philip, ``An
  unforgeable publicly verifiable watermark for large language models,'' in
  \emph{The Twelfth International Conference on Learning Representations},
  2023.

\bibitem[Wu et~al.()Wu, Hu, Guo, Zhang, and Huang]{wu2023dipmark}
Y.~Wu, Z.~Hu, J.~Guo, H.~Zhang, and H.~Huang, ``A resilient and accessible
  distribution-preserving watermark for large language models,'' in
  \emph{Forty-first International Conference on Machine Learning}.

\bibitem[Hu et~al.()Hu, Chen, Wu, Wu, Zhang, and Huang]{huunbiased}
Z.~Hu, L.~Chen, X.~Wu, Y.~Wu, H.~Zhang, and H.~Huang, ``Unbiased watermark for
  large language models,'' in \emph{The Twelfth International Conference on
  Learning Representations}.

\bibitem[Fairoze et~al.(2023)Fairoze, Garg, Jha, Mahloujifar, Mahmoody, and
  Wang]{fairoze2023publicly}
J.~Fairoze, S.~Garg, S.~Jha, S.~Mahloujifar, M.~Mahmoody, and M.~Wang,
  ``Publicly detectable watermarking for language models,'' \emph{Cryptology
  ePrint Archive}, 2023.

\bibitem[Christ and Gunn(2024)]{christ2024pseudorandom}
M.~Christ and S.~Gunn, ``Pseudorandom error-correcting codes,'' in \emph{Annual
  International Cryptology Conference}.\hskip 1em plus 0.5em minus 0.4em\relax
  Springer, 2024, pp. 325--347.

\bibitem[Ghentiyala and Guruswami(2024)]{ghentiyala2024new}
S.~Ghentiyala and V.~Guruswami, ``New constructions of pseudorandom codes,''
  \emph{Cryptology ePrint Archive}, 2024.

\bibitem[Golowich and Moitra()]{golowichedit}
N.~Golowich and A.~Moitra, ``Edit distance robust watermarks via indexing
  pseudorandom codes,'' in \emph{The Thirty-eighth Annual Conference on Neural
  Information Processing Systems}.

\bibitem[Zhao et~al.(2024{\natexlab{a}})Zhao, Gunn, Christ, Fairoze, Fabrega,
  Carlini, Garg, Hong, Nasr, Tramer, et~al.]{zhao2024sok}
X.~Zhao, S.~Gunn, M.~Christ, J.~Fairoze, A.~Fabrega, N.~Carlini, S.~Garg,
  S.~Hong, M.~Nasr, F.~Tramer \emph{et~al.}, ``Sok: Watermarking for
  ai-generated content,'' \emph{arXiv preprint arXiv:2411.18479}, 2024.

\bibitem[Gabrilovich and Gontmakher(2002)]{gabrilovich2002homograph}
E.~Gabrilovich and A.~Gontmakher, ``The homograph attack,''
  \emph{Communications of the ACM}, vol.~45, no.~2, p. 128, 2002.

\bibitem[Helfrich and Neff(2012)]{helfrich2012dual}
J.~N. Helfrich and R.~Neff, ``Dual canonicalization: An answer to the homograph
  attack,'' in \emph{2012 eCrime Researchers Summit}.\hskip 1em plus 0.5em
  minus 0.4em\relax IEEE, 2012, pp. 1--10.

\bibitem[Pajola and Conti(2021)]{pajola2021fall}
L.~Pajola and M.~Conti, ``Fall of giants: How popular text-based mlaas fall
  against a simple evasion attack,'' in \emph{2021 IEEE European Symposium on
  Security and Privacy (EuroS\&P)}.\hskip 1em plus 0.5em minus 0.4em\relax
  IEEE, 2021, pp. 198--211.

\bibitem[Boucher et~al.(2022)Boucher, Shumailov, Anderson, and
  Papernot]{boucher2022bad}
N.~Boucher, I.~Shumailov, R.~Anderson, and N.~Papernot, ``Bad characters:
  Imperceptible nlp attacks,'' in \emph{2022 IEEE Symposium on Security and
  Privacy (SP)}.\hskip 1em plus 0.5em minus 0.4em\relax IEEE, 2022, pp.
  1987--2004.

\bibitem[Goodside(2023)]{goodside2023adversarial}
R.~Goodside, ``There are adversarial attacks for that proposal as well --- in
  particular, generating with emojis after words and then removing them before
  submitting defeats it,'' Twitter, Jan. 2023, uRL:
  \url{https://twitter.com/goodside/status/1610682909647671306}.

\bibitem[Sadasivan et~al.(2023)Sadasivan, Kumar, Balasubramanian, Wang, and
  Feizi]{sadasivan2023can}
V.~S. Sadasivan, A.~Kumar, S.~Balasubramanian, W.~Wang, and S.~Feizi, ``Can
  ai-generated text be reliably detected?'' \emph{arXiv preprint
  arXiv:2303.11156}, 2023.

\bibitem[Krishna et~al.(2023)Krishna, Song, Karpinska, Wieting, and
  Iyyer]{krishna2023paraphrasing}
K.~Krishna, Y.~Song, M.~Karpinska, J.~Wieting, and M.~Iyyer, ``Paraphrasing
  evades detectors of ai-generated text, but retrieval is an effective
  defense,'' \emph{arXiv preprint arXiv:2303.13408}, 2023.

\bibitem[Piet et~al.(2023)Piet, Sitawarin, Fang, Mu, and Wagner]{piet2023mark}
J.~Piet, C.~Sitawarin, V.~Fang, N.~Mu, and D.~Wagner, ``Mark my words:
  Analyzing and evaluating language model watermarks,'' \emph{arXiv preprint
  arXiv:2312.00273}, 2023.

\bibitem[Jovanovi{\'c} et~al.(2024)Jovanovi{\'c}, Staab, and
  Vechev]{jovanovic2024watermark}
N.~Jovanovi{\'c}, R.~Staab, and M.~Vechev, ``Watermark stealing in large
  language models,'' \emph{arXiv preprint arXiv:2402.19361}, 2024.

\bibitem[Zhang et~al.(2024)Zhang, Edelman, Francati, Venturi, Ateniese, and
  Barak]{zhangwatermarks2024}
H.~Zhang, B.~L. Edelman, D.~Francati, D.~Venturi, G.~Ateniese, and B.~Barak,
  ``Watermarks in the sand: Impossibility of strong watermarking for language
  models,'' in \emph{Forty-first International Conference on Machine Learning},
  2024.

\bibitem[OpenAI(2023)]{openai2023gpt}
R.~OpenAI, ``Gpt-4 technical report. arxiv 2303.08774,'' \emph{View in
  Article}, vol.~2, no.~5, 2023.

\bibitem[Touvron et~al.(2023)Touvron, Martin, Stone, Albert, Almahairi, Babaei,
  Bashlykov, Batra, Bhargava, Bhosale, et~al.]{touvron2023llama}
H.~Touvron, L.~Martin, K.~Stone, P.~Albert, A.~Almahairi, Y.~Babaei,
  N.~Bashlykov, S.~Batra, P.~Bhargava, S.~Bhosale \emph{et~al.}, ``Llama 2:
  Open foundation and fine-tuned chat models,'' \emph{arXiv preprint
  arXiv:2307.09288}, 2023.

\bibitem[Zhao et~al.(2023)Zhao, Ananth, Li, and Wang]{zhao2023provable}
X.~Zhao, P.~Ananth, L.~Li, and Y.-X. Wang, ``Provable robust watermarking for
  ai-generated text,'' \emph{arXiv preprint arXiv:2306.17439}, 2023.

\bibitem[Lu et~al.(2024)Lu, Liu, Yu, Li, and King]{lu2024entropy}
Y.~Lu, A.~Liu, D.~Yu, J.~Li, and I.~King, ``An entropy-based text watermarking
  detection method,'' \emph{arXiv preprint arXiv:2403.13485}, 2024.

\bibitem[Pan et~al.(2024)Pan, Liu, He, Gao, Zhao, Lu, Zhou, Liu, Hu, Wen,
  et~al.]{pan2024markllm}
L.~Pan, A.~Liu, Z.~He, Z.~Gao, X.~Zhao, Y.~Lu, B.~Zhou, S.~Liu, X.~Hu, L.~Wen
  \emph{et~al.}, ``Markllm: An open-source toolkit for llm watermarking,''
  \emph{arXiv preprint arXiv:2405.10051}, 2024.

\bibitem[Raffel et~al.(2020)Raffel, Shazeer, Roberts, Lee, Narang, Matena,
  Zhou, Li, and Liu]{raffel2020exploring}
C.~Raffel, N.~Shazeer, A.~Roberts, K.~Lee, S.~Narang, M.~Matena, Y.~Zhou,
  W.~Li, and P.~J. Liu, ``Exploring the limits of transfer learning with a
  unified text-to-text transformer,'' \emph{Journal of machine learning
  research}, vol.~21, no. 140, pp. 1--67, 2020.

\bibitem[He et~al.(2024)He, Zhou, Hao, Liu, Wang, Tu, Zhang, and
  Wang]{hewatermarks2024}
\BIBentryALTinterwordspacing
Z.~He, B.~Zhou, H.~Hao, A.~Liu, X.~Wang, Z.~Tu, Z.~Zhang, and R.~Wang, ``Can
  watermarks survive translation? on the cross-lingual consistency of text
  watermark for large language models,'' in \emph{Proceedings of the 62nd
  Annual Meeting of the Association for Computational Linguistics (Volume 1:
  Long Papers)}, L.-W. Ku, A.~Martins, and V.~Srikumar, Eds.\hskip 1em plus
  0.5em minus 0.4em\relax Bangkok, Thailand: Association for Computational
  Linguistics, Aug. 2024, pp. 4115--4129. [Online]. Available:
  \url{https://aclanthology.org/2024.acl-long.226}
\BIBentrySTDinterwordspacing

\bibitem[Huo et~al.()Huo, Somayajula, Liang, Zhang, Koushanfar, and
  Xie]{huotoken}
M.~Huo, S.~A. Somayajula, Y.~Liang, R.~Zhang, F.~Koushanfar, and P.~Xie,
  ``Token-specific watermarking with enhanced detectability and semantic
  coherence for large language models,'' in \emph{Forty-first International
  Conference on Machine Learning}.

\bibitem[Zhou et~al.(2024)Zhou, Zhao, Xu, and Ren]{zhou2024bileve}
T.~Zhou, X.~Zhao, X.~Xu, and S.~Ren, ``Bileve: Securing text provenance in
  large language models against spoofing with bi-level signature,'' \emph{arXiv
  preprint arXiv:2406.01946}, 2024.

\bibitem[Liu et~al.(2024)Liu, Pan, Hu, Meng, and Wen]{liu2024a}
\BIBentryALTinterwordspacing
A.~Liu, L.~Pan, X.~Hu, S.~Meng, and L.~Wen, ``A semantic invariant robust
  watermark for large language models,'' in \emph{The Twelfth International
  Conference on Learning Representations}, 2024. [Online]. Available:
  \url{https://openreview.net/forum?id=6p8lpe4MNf}
\BIBentrySTDinterwordspacing

\bibitem[Zhao et~al.(2024{\natexlab{b}})Zhao, Li, and Wang]{zhao2024permute}
X.~Zhao, L.~Li, and Y.-X. Wang, ``Permute-and-flip: An optimally robust and
  watermarkable decoder for llms,'' \emph{arXiv preprint arXiv:2402.05864},
  2024.

\bibitem[Fu et~al.(2024)Fu, Zhao, Yang, Zhang, Chen, and
  Xiao]{fu2024gumbelsoft}
J.~Fu, X.~Zhao, R.~Yang, Y.~Zhang, J.~Chen, and Y.~Xiao, ``Gumbelsoft:
  Diversified language model watermarking via the gumbelmax-trick,''
  \emph{arXiv preprint arXiv:2402.12948}, 2024.

\bibitem[Cohen et~al.(2024)Cohen, Hoover, and
  Schoenbach]{cohen2024watermarking}
A.~Cohen, A.~Hoover, and G.~Schoenbach, ``Watermarking language models for many
  adaptive users,'' in \emph{2025 IEEE Symposium on Security and Privacy
  (SP)}.\hskip 1em plus 0.5em minus 0.4em\relax IEEE Computer Society, 2024,
  pp. 84--84.

\bibitem[Welleck et~al.()Welleck, Kulikov, Roller, Dinan, Cho, and
  Weston]{welleckneural}
S.~Welleck, I.~Kulikov, S.~Roller, E.~Dinan, K.~Cho, and J.~Weston, ``Neural
  text generation with unlikelihood training,'' in \emph{International
  Conference on Learning Representations}.

\bibitem[Li et~al.(2022)Li, Holtzman, Fried, Liang, Eisner, Hashimoto,
  Zettlemoyer, and Lewis]{li2022contrastive}
X.~L. Li, A.~Holtzman, D.~Fried, P.~Liang, J.~Eisner, T.~Hashimoto,
  L.~Zettlemoyer, and M.~Lewis, ``Contrastive decoding: Open-ended text
  generation as optimization,'' \emph{arXiv preprint arXiv:2210.15097}, 2022.

\bibitem[Wieting et~al.(2022)Wieting, Gimpel, Neubig, and
  Berg-Kirkpatrick]{wieting2022paraphrastic}
J.~Wieting, K.~Gimpel, G.~Neubig, and T.~Berg-Kirkpatrick, ``Paraphrastic
  representations at scale,'' in \emph{Proceedings of the 2022 Conference on
  Empirical Methods in Natural Language Processing: System Demonstrations},
  2022, pp. 379--388.

\bibitem[{European Commission}(2021)]{euaiact}
{European Commission}, ``{The EU Artificial Intelligence Act},'' Available at:
  \url{https://artificialintelligenceact.eu/}, 2021, proposed regulation
  focusing on transparency and accountability in high-risk AI systems.

\bibitem[{OECD}(2019)]{oecdaiprinciples}
{OECD}, ``Oecd ai principles,'' Available at:
  \url{https://oecd.ai/en/dashboards/ai-principles/}, 2019, guidelines for
  ethical, trustworthy AI. Transparency and accountability are key principles.

\bibitem[{National Institute of Standards and Technology
  (NIST)}(2023)]{nistframework}
{National Institute of Standards and Technology (NIST)}, ``Ai risk management
  framework,'' U.S. Department of Commerce, Tech. Rep., 2023, framework to
  improve AI system trustworthiness and manage risks, emphasizing transparency.

\bibitem[Brown et~al.(2020)Brown, Mann, Ryder, Subbiah, Kaplan, Dhariwal,
  Neelakantan, Ouyang, and Amodei]{openairesearch}
T.~Brown, B.~Mann, N.~Ryder, D.~Subbiah, J.~Kaplan, P.~Dhariwal,
  A.~Neelakantan, L.~Ouyang, and D.~Amodei, ``Language models are few-shot
  learners,'' \emph{Advances in Neural Information Processing Systems}, 2020,
  seminal paper introducing GPT-3, discussing the importance of probabilities
  in understanding model behavior.

\bibitem[{OpenAI}(2023)]{openaiapidocs}
{OpenAI}, ``Openai api documentation,'' Available at:
  \url{https://platform.openai.com/docs/}, 2023, developer documentation
  highlighting the use of top-K sampling, beam search, and probability outputs.

\end{thebibliography}
